\documentclass[11pt, a4paper]{article}

\usepackage[utf8]{inputenc}
\usepackage[T1]{fontenc}
\usepackage{lmodern}
\usepackage{amsmath,amssymb}
\usepackage{booktabs,array,tabularx,longtable}
\usepackage{graphicx,adjustbox,threeparttable}
\usepackage{geometry,microtype,titlesec,enumitem,placeins}
\usepackage{natbib}
\usepackage{hyperref}
\newcolumntype{P}[1]{>{\raggedright\arraybackslash}p{#1}}

\titleformat{\section}{\large\bfseries\sffamily}{\thesection}{1em}{}
\titleformat{\subsection}{\normalsize\bfseries\sffamily}{\thesubsection}{1em}{}

\hypersetup{
    colorlinks=true,
    linkcolor=blue,
    citecolor=blue,
    filecolor=magenta,      
    urlcolor=cyan,
    pdftitle={How Do Agent Harnesses Create Value? Planning Information and Release Control in Stateful LLM Agents},
    pdfauthor={Yukun Zhang, Kemu Xu, Yishen Chen},
}

\title{How Do Agent Harnesses Create Value? Planning Information and Release Control in Stateful LLM Agents}

\author{\begin{tabular}{@{}c@{\hspace{1em}}c@{\hspace{1em}}c@{}}
\normalsize Yukun Zhang & \normalsize Kemu Xu & \normalsize Yishen Chen\\
\small The Chinese University & \small University of Edinburgh & \small The Chinese University\\
\small of Hong Kong & & \small of Hong Kong, Shenzhen\\
\small Hong Kong, China & \small Edinburgh, United Kingdom & \small Shenzhen, China\\
\footnotesize\texttt{215010026@link.cuhk.edu.cn} &
\footnotesize\texttt{s2749200@ed.ac.uk} &
\footnotesize\texttt{yishenchen@link.cuhk.edu.cn}
\end{tabular}}
\date{}

\begin{document}

\maketitle

\begin{abstract}
Agent harnesses supply planning guidance, organize execution, and check
completion. We study how these components affect success, erroneous
acceptance, and cost in two Retail experiments and an Airline pilot in
$\tau^2$-bench. The primary comparison pairs prewritten task-specific plans
(Fixed) with shuffled policy text matched in word count (Sham), isolating
the contribution of guidance content. Across 265 matched cells,
Fixed improves oracle-verified success by 7.17 percentage points
(90\% task-clustered bootstrap interval, 1.15--13.36 points), with gains
concentrated in higher-complexity tasks. A read-only terminal verifier
rejects 61\% of Retail oracle-invalid episodes while withholding 17\% of correct
ones, at less than one cent of additional cost per episode.
Which component matters more depends on the loss assigned to
erroneous acceptance: at low liability the planning gain dominates;
at high liability the verifier's avoided false passes dominate---and
a standalone verifier captures nearly all the false-pass benefit of the
full planning-plus-verification stack at a fraction of its cost.
\end{abstract}

\section{Introduction}
\label{sec:introduction}

\subsection{The Attribution Problem in Agent Performance}
\label{subsec:intro_attribution}

Large language model agents retrieve information, invoke tools, and modify
persistent environment states. Their outcomes depend on both the model and
the surrounding mechanisms that organize its work: planning, repair, memory,
tool orchestration, and completion checks. We call this execution and control
structure an \emph{agent harness}. Reasoning--action integration, reflective
feedback, and memory management illustrate how external organization shapes
the behavior of a given model
\citep{yao2023react,shinn2023reflexion,packer2023memgpt}.

Interactive benchmarks and system-level comparisons measure what models and
harnesses achieve together
\citep{liu2024agentbench,ma2024agentboard,yao2025taubench,kapoor2025hal}.
Attributing that performance to a component requires a more specific
comparison. Additional guidance, inference expenditure, execution constraints,
and repair opportunities can all change success. A useful component
evaluation states what support changes, which outcome it affects, and what
resources it consumes.

\subsection{Planning Information, Release Control, and Operating Cost}
\label{subsec:intro_channels}

Planning supplies subgoals, dependencies, and execution guidance that an
executor may find useful
\citep{zhou2023least,wang2023plansolve,erdogan2025planact}.
Comparing a supplied plan with Minimal changes both the guidance and the
amount of context. A control matched in word count and packaging helps
isolate the contribution of the supplied task-specific content.

Verification addresses acceptance of an attempted completion---\emph{release
control} in the language of the title. A terminal verifier
can withhold an invalid result, but can also reject a correct one.
Oracle correctness, a completion claim, and the acceptance decision therefore
need separate measures. This distinction is especially consequential in
stateful environments: a terminal rejection occurs after execution and may
leave earlier refunds, cancellations, or other state changes in place.

Both components consume resources. Their practical value depends on the
benefit assigned to successful completion, the loss assigned to erroneous
acceptance, and the cost of obtaining and handling the result. We use these
three quantities to connect component outcomes to conditional configuration
value.

\subsection{Research Questions and Experimental Approach}
\label{subsec:intro_design}

We ask whether supplied planning guidance improves execution relative to a
specified context control, how terminal verification changes acceptance outcomes,
and how outcome valuations affect configuration comparisons.
The study uses Retail and Airline from $\tau^2$-bench
\citep{barres2025tau2}. Shared Retail contributes 1,547 trajectories from six
models, sixteen tasks, and seven configurations. A separate planner-focused
Retail experiment contributes 1,227 trajectories from five models,
twenty-four tasks, and four planning conditions. Airline contributes 233
trajectories from five models, six tasks, and four configurations.

The primary comparison assigns prewritten task-specific Fixed guidance or
Sham text formed from shuffled policy words, matched in whitespace-word count
and input wrapper. Both use static text. We pair records by model, task, and
replication and quantify uncertainty by task-clustered bootstrap.

Verification uses pooled descriptive rates, model-specific matched contrasts,
and direct rejection counts. The value exercise pairs each configuration
with its Minimal counterparts and applies explicit success values
and false-pass liabilities.

\subsection{Main Findings and Contributions}
\label{subsec:intro_findings}

Fixed improves oracle-verified success over Sham by 7.17 percentage points
(90\% interval, 1.15--13.36 points), with gains concentrated in
higher-complexity tasks. DeepSeek Flash and Qwen show the largest
effects; Kimi's estimate is near zero despite a similar baseline. All
leave-one-model-out point estimates remain positive, and the Sham--Minimal
estimate (1.89 points) provides no evidence that the shuffled control
harms performance.

In shared Retail, the terminal verifier rejects 83 of 137 oracle-invalid
episodes (61\%) while withholding 16 of 92 correct ones (17\%), reducing the
false-pass rate from 57.21\% to 20.96\%. Matched contrasts are negative
for all six Retail models. Verifier-only captures nearly the same
avoided false-pass benefit as the full stack (48.4 vs.\ 49.8 pp) at a
twelfth of the incremental cost.

The paper contributes:
\begin{enumerate}[leftmargin=*, itemsep=3pt, topsep=4pt]
\item A matched Fixed--Sham comparison of prewritten task-specific guidance
against a word-count-matched context control.
\item An outcome framework and rejection analysis that separate oracle
correctness from terminal acceptance.
\item A conditional value framework combining success, avoided false pass,
and inference cost under stated liability weights.
\end{enumerate}

Section~\ref{sec:related-literature} reviews related work.
Sections~\ref{sec:conceptual_framework} and~\ref{sec:experimental_design}
define the framework and design. Sections~\ref{sec:planner_results}
through~\ref{sec:liability_analysis} present planning, verification, and
scenario results, followed by discussion and conclusion.

\section{Related Work}
\label{sec:related-literature}

\paragraph{Agent evaluation beyond scalar scores.}
Model benchmarks measure standardized knowledge and code generation
\citep{hendrycks2021mmlu,chen2021evaluating}, while HELM evaluates multiple
dimensions, including accuracy, calibration, robustness, and efficiency
\citep{liang2023helm}. Agent evaluation extends to general-assistant questions
requiring reasoning, multimodal understanding, browsing, and tool use
\citep{mialon2023gaia}, as well as decision-making across interactive
environments \citep{liu2024agentbench}. AgentBoard measures incremental
progress during multi-turn interaction \citep{ma2024agentboard}; WebArena and
OSWorld assess functional completion in web and computer environments
\citep{zhou2023webarena,xie2024osworld}. These approaches already provide
information beyond a single model score. Our focus is the contribution of
specific harness interventions to verified success, false passes, and
operating cost within common model--task conditions.

\paragraph{Tool use, software work, and long-horizon evaluation.}
ToolLLM combines data construction, model training, API retrieval, multi-tool
execution, and evaluation \citep{qin2024toolllm}. BFCL tests serial and
parallel function calls, abstention, and stateful multi-step behavior
\citep{patil2025bfcl}. For tool--agent--user interaction, $\tau$-bench provides
domain policies and evaluates terminal database states and required response
content; its success reward does not ensure full policy compliance
\citep{yao2025taubench}. Software-engineering benchmarks
use executable tests or task values grounded in freelance payments
\citep{jimenez2024swebench,miserendino2025swelancer}. RE-Bench compares agents
with human research engineers under time budgets \citep{wijk2025rebench},
while time-horizon evaluation relates agent success to the duration of tasks
for human workers \citep{kwa2025longtasks}. HAL jointly analyzes models,
scaffolds, and benchmarks \citep{kapoor2025hal}. We complement this evaluation
literature with a controlled contrast of planning information and
separate comparisons of verification configurations.

\paragraph{Planning information and inference-time organization.}
Reasoning methods expose intermediate steps, aggregate answers across sampled
paths, or decompose problems into ordered subproblems
\citep{wei2022chain,wang2023selfconsistency,zhou2023least,wang2023plansolve}.
ReAct interleaves reasoning and actions \citep{yao2023react}, whereas Toolformer
learns tool invocation through training \citep{schick2023toolformer}.
Tree of Thoughts, RAP, and LATS search candidate reasoning or action paths
\citep{yao2023tree,hao2023rap,zhou2024lats}; DEPS revises plans using execution
feedback \citep{wang2023deps}. Plan-and-Act explicitly separates a high-level
planner from an environment-specific executor and trains the planner using
synthetic data \citep{erdogan2025planact}. PlanBench tests plan construction
and reasoning about change \citep{valmeekam2023planbench}, while APB diagnoses
planning under extraneous tools, broken tools, and infeasible tasks
\citep{sun2026apb}. Test-time scaling studies examine how difficulty, model
size, inference strategy, and compute budget affect reasoning performance
\citep{snell2024scaling,wu2025inference}.

Prompt interventions also examine which information drives planning gains.
\citet{verma2024brittle} vary ReAct exemplars and placebo guidance to study
sensitivity to content and exemplar--query similarity. Our Fixed--Sham
comparison extends this attribution question to stateful tool use with
prewritten task-specific guidance, a word-count-matched context control,
and paired oracle outcomes.

\paragraph{Verification signals and terminal completion.}
Outcome verifiers rank candidate solutions, while process-based feedback
evaluates intermediate reasoning
\citep{cobbe2021training,uesato2022process,lightman2023verify}.
Self-verification supplies scores for candidate-answer selection
\citep{weng2023selfverification}. Model-based judges assess open-ended text
\citep{zheng2023judging,liu2023geval}; Prometheus conditions evaluation on
rubrics and reference answers \citep{kim2024prometheus}, and RewardBench tests
reward-model preference ranking \citep{lambert2025rewardbench}. Repeated
sampling expands candidate coverage \citep{brown2024large}, while adaptive
computation allocates search and revision effort \citep{snell2024scaling}.
Verification signals can guide selection, search, or training
\citep{snell2024scaling,setlur2025verification}. These uses of verification
differ from deciding whether an agent's claimed terminal state warrants
release.

Recent agent studies address this terminal distinction directly. VIGIL scores
world-state completion separately from the correctness of an agent's terminal
report \citep{chen2026vigil}. \citet{advani2026falsesuccess} studies false success,
where completion claims conflict with state-based ground truth, and evaluates
detectors on $\tau^2$-bench and AppWorld. Procedure-Aware Evaluation instead
examines corrupt success: positive benchmark outcomes accompanied by
procedural violations \citep{cao2026procedure}. A corrupt success need not be
a false pass under our terminal-objective oracle. Computer-use verification
research also separates process and outcome scoring and evaluates trajectory
judges against human labels \citep{rosset2026verifiers}. Our contribution
builds on these distinctions by measuring how external verification relates
to verified success and false-pass control, then valuing the release
margin under downstream liability.

\paragraph{Harness runtime, risk, and cost-sensitive evaluation.}
Harness research already treats runtime support as an object of controlled
comparison. \citet{zhong2026harness} propose an H0--H3 ladder that varies
support while holding the model and task fixed, with an illustrative
validation case and a taxonomy separating task outcomes from verification
evidence. SGLang shows how runtime design affects the efficiency of structured
language-model programs \citep{zheng2024sglang}. At the evaluation level,
\citet{kapoor2024agentsmatter} argue for cost-controlled agent comparisons and
joint accuracy--cost optimization; FrugalGPT studies budget-aware model
cascades \citep{chen2024frugalgpt}. These works motivate assessing the value of
additional support relative to its resource demands.

Risk-sensitive evaluation adds a further dimension. ToolEmu tests long-tail
risks in emulated tool environments \citep{ruan2024toolemu}, and AgentDojo
evaluates task utility alongside prompt-injection robustness
\citep{debenedetti2024agentdojo}. The Verifier Tax compares tool-calling,
planning, and policy-mediated architectures in Retail and Airline domains,
documenting safety--success tradeoffs and verification overhead
\citep{sah2026verifiertax}. Selective prediction provides a
related risk--coverage perspective through an explicit reject option
\citep{geifman2019selectivenet}. We connect these perspectives to component
comparisons and a scenario analysis of verified completion, avoided false
passes, and incremental execution cost.

\section{Evaluation Framework: Planning Information, Verification,
and Conditional Harness Value}
\label{sec:conceptual_framework}

An agent harness can affect task outcomes by changing the information
available to the executor, organizing its interaction with tools,
or evaluating a proposed completion. These interventions need not
improve the same outcome dimension. Planning may help an executor
complete a task, while verification may reduce the frequency with
which an incorrect outcome is treated as complete. Both interventions
also consume resources.

\begin{figure}[t]
\centering
\includegraphics[width=\linewidth]{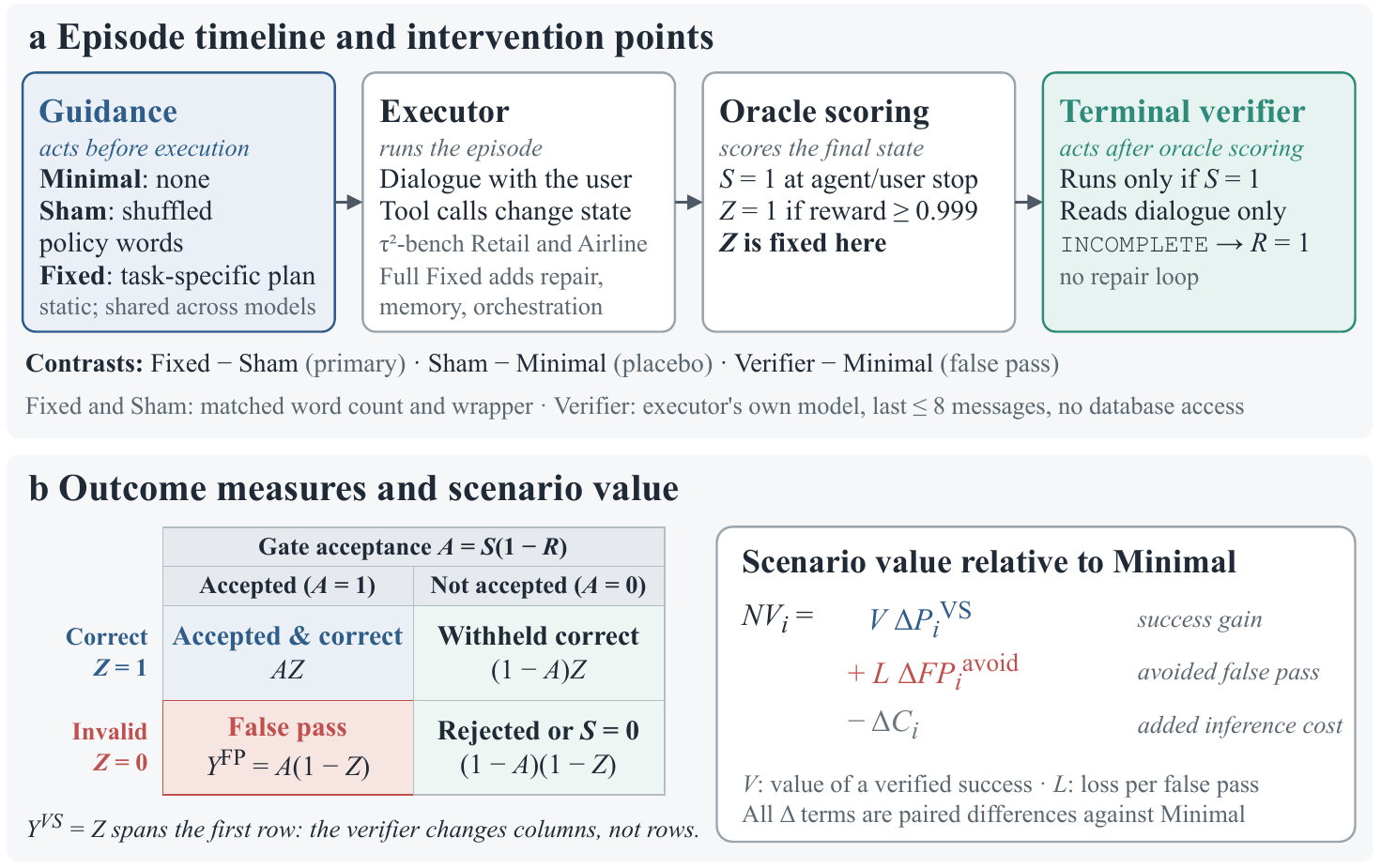}
\caption{\textbf{Planning guidance, terminal verification, and scenario value.}
(a) Minimal supplies no guidance; Fixed supplies a task-specific plan; Sham
supplies shuffled policy text matched to Fixed in word count and wrapper.
Oracle scoring fixes \(Z\) before the terminal verifier sets rejection flag
\(R\) from recent dialogue. (b) The outcome matrix separates oracle
correctness from acceptance \(A=S(1-R)\): verified success is \(Y^{\mathrm{VS}}=Z\),
while false pass is \(Y^{\mathrm{FP}}=A(1-Z)\). The value card combines
paired changes in success, avoided false passes, and logged inference cost
relative to Minimal, as in Equation~\ref{eq:component_net_value}.}
\label{fig:harness_value_framework}
\end{figure}

We organize the analysis around oracle success, false pass, and logged
inference cost. Figure~\ref{fig:harness_value_framework} connects the
guidance intervention, post-hoc check, and outcome measures.

\subsection{Harness Components and Their Intervention Points}
\label{subsec:framework_intervention_points}

The unit of observation is an execution trajectory indexed by
\((m,t,h,r)\), where \(m\) denotes the model, \(t\) the task,
\(h\) the harness configuration, and \(r\) the replication.
The model interacts with a stateful environment under the execution
protocol associated with \(h\). The resulting trajectory produces
an evaluated task outcome and a record of resource consumption.

Harness components enter this process at different points.
Planning supplies task structure, including subgoals, dependencies,
and execution guidance. Tool orchestration constrains or organizes
tool interactions. Memory preserves information across steps.
Evaluation and repair provide opportunities to inspect outcomes
and revise behavior. Terminal verification reviews a proposed
completion before the system accepts it.

Component effects can overlap: planning may prevent errors that would
otherwise trigger evaluation, and several checks may consume the same
execution budget. We evaluate targeted component contrasts and treat
Full Fixed as the combined outcome of its implemented components.

\subsection{Observed Outcomes and Resource Use}
\label{subsec:framework_observed_outcomes}

We distinguish oracle correctness, episode termination, and verifier rejection.
Let \(Z=1\) when the benchmark reward is at least \(0.999\),
\(S=1\) when termination is \texttt{USER\_STOP} or \texttt{AGENT\_STOP},
and \(R=1\) when the terminal verifier rejects the result.
The two quality measures are
\begin{equation}
Y^{\mathrm{VS}}=Z,\qquad
Y^{\mathrm{FP}}=S(1-R)(1-Z).
\label{eq:observed_outcomes}
\end{equation}
The logged \texttt{agent\_asserted} field supplies \(S\), a termination-based
proxy for completion. The gate-acceptance indicator derived from these
fields is \(A=S(1-R)\); accepted and oracle-correct completion is \(AZ\).
Verified success counts oracle-correct episodes, including those subsequently
rejected by the verifier. Appendix~\ref{app:verifier_results} reports the
rejection cross-tabulation.

Resource measures include tokens, tool calls, simulation latency, and logged
execution cost. Cost@Success and Pass@Budget describe different aspects of
this record.

\subsection{Planning Information and Model--Task Heterogeneity}
\label{subsec:framework_planning_information}

The planning comparison evaluates supplied task-specific guidance.
Fixed provides a prewritten plan; Sham supplies shuffled domain-policy
words matched to its whitespace-word count and common wrapper. The
sample-matched estimator is defined in
Equation~\ref{eq:planner_primary_estimand}.

This contrast holds the model, task, and replication key fixed while
changing the added guidance.
Sham--Minimal measures the behavior of the chosen context control,
and Self contrasts assess guidance generated by the executor.

Planning returns can vary with both task demands and the executor's
ability to implement a plan. Baseline success and reference-action
complexity provide descriptive dimensions for examining that variation.

\subsection{Verification and the Interpretation of False Pass}
\label{subsec:framework_verification}

The evaluated verifier runs after task execution. For episodes with \(S=1\),
it reviews recent dialogue and sets the rejection indicator \(R\).
This operation directly affects the false-pass measure in
Equation~\ref{eq:observed_outcomes}; the oracle outcome \(Z\) is already fixed.
The contrast between these outcome margins motivates separate reporting of
oracle success, false pass, and rejection.

A \emph{pseudo-completion} is an episode with false pass equal to one;
Section~\ref{sec:verifier_results} uses this label.

\subsection{Conditional Value under Cost and Risk}
\label{subsec:framework_conditional_value}

The same outcome difference can have different practical value
across deployments. A successful completion may be highly valuable
in one application, while preventing an erroneous completion
may dominate the decision in another. Execution cost provides
a further source of variation.

To express these trade-offs, consider configuration \(i\)
relative to Minimal. Let
\(\Delta P_i^{\mathrm{VS}}\) denote its difference
in verified-success probability,
\(\Delta FP_i^{\mathrm{avoid}}\) its reduction
in false-pass probability, and
\(\Delta C_i\) its incremental logged execution cost.
For a scenario assigning value \(V\) to a verified completion
and loss \(L\) to a false pass, incremental net value is
\begin{equation}
NV_i(V,L)
=
V\Delta P_i^{\mathrm{VS}}
+
L\Delta FP_i^{\mathrm{avoid}}
-
\Delta C_i.
\label{eq:component_net_value}
\end{equation}

The first term values additional oracle-success outcomes.
The second term values avoided false passes.
The third accounts for additional execution expenditure.
This decomposition permits a configuration to have positive
scenario value even when its verified-success difference is
small or negative, provided that its reduction in false pass
receives sufficient weight.

For a configuration with
\(\Delta FP_i^{\mathrm{avoid}}>0\), the corresponding
break-even false-pass loss is
\begin{equation}
L_i^*(V)
=
\frac{
\Delta C_i-V\Delta P_i^{\mathrm{VS}}
}{
\Delta FP_i^{\mathrm{avoid}}
}.
\label{eq:framework_break_even}
\end{equation}
Under the aggregate differences, the configuration
has non-negative incremental scenario value when
\(L\geq L_i^*(V)\). A negative threshold means that its
combination of verified-success value and logged execution
cost is already favorable relative to Minimal at \(L=0\).

The threshold compares a configuration with Minimal.
The weights \(V\) and \(L\) are scenario inputs.

\section{Experimental Design, Data, and Evidence Scope}
\label{sec:experimental_design}

The empirical study uses the Retail and Airline environments supplied by
$\tau^2$-bench \citep{barres2025tau2}. Retail provides the primary evidence through
the shared Retail experiment and a separate planner-focused
Retail experiment. Airline provides a smaller evaluation in a second
environment. We describe the implemented interventions, scoring rules, matching,
missingness, and uncertainty procedures below.

\subsection{Evaluation Environments and Task Coverage}
\label{subsec:design_environments}

In both environments, an agent receives a task instruction and
interacts with structured tools that expose information or modify
an underlying state. Task completion is evaluated by a hidden
oracle independently of the executor's own completion claim.
Successful performance therefore requires an acceptable task
outcome under the environment's evaluation criteria, rather than
a persuasive statement that the task has been completed.

Table~\ref{tab:design_realized_coverage} summarizes realized coverage
of the three trajectory blocks; missing cells are detailed below.

\begin{table}[t]
\centering
\small
\caption{Realized coverage of the trajectory experiments.}
\label{tab:design_realized_coverage}
\setlength{\tabcolsep}{6pt}
\begin{tabular}{@{}lrrrrr@{}}
\toprule
\textbf{Experimental block}
& \textbf{Models}
& \textbf{Tasks}
& \textbf{Configs}
& \textbf{Reps}
& \textbf{Trajectories}
\\
\midrule
Retail shared
& 6 & 16 & 7 & \(\leq 3\) & 1,547
\\
Retail planner
& 5 & 24 & 4 & \(\leq 3\) & 1,227
\\
Airline pilot
& 5 & 6 & 4 & 2 & 233
\\
\bottomrule
\end{tabular}
\end{table}

\subsubsection{Retail: Shared Experiment}
\label{subsubsec:design_retail_shared}

Retail represents customer-service workflows involving order
modification, returns and exchanges, payment disputes, account
operations, shipment tracking, and multi-step fulfillment.
Depending on the task, execution may require authentication,
eligibility checks, information retrieval, correctly ordered
tool calls, confirmation, or recovery from an intermediate
failure. Some actions modify persistent state, while others
depend on preconditions established by earlier interactions.

The shared Retail experiment contains 1,547 valid trajectories
from six models, sixteen tasks, and seven harness configurations,
with up to three replications per model--task--configuration cell.
It supplies the configuration-level performance and cost results,
the supplementary shared-Retail planning contrasts, and the
trajectory outcomes used in the verifier and capability-conditioned
analyses.

The supplementary planning comparison, FDS-by-verifier analysis,
liability scenarios, and incremental prediction analysis
reuse these shared trajectories. The independent capability probes
and tool-coordination stress probes provide separate measurement
data. Appendix~\ref{app:protocol} maps all experiment identifiers to their
data sources.

\subsubsection{Retail: Planner-Focused Experiment}
\label{subsubsec:design_retail_planner}

The planner-focused Retail experiment contains 1,227 valid
trajectories from five models, twenty-four tasks, and four
conditions: Minimal, Planner Fixed, Planner Sham, and Planner
Self. Each model--task--condition cell has up to three
replications. This block supplies the primary Fixed--Sham
comparison and the associated secondary planning analyses.

Section~\ref{subsec:design_samples} defines the primary analysis sample;
the model-specific results also retain the fifth model.

Task structure is summarized from reference actions: substeps count the
actions, the dependency-depth proxy is \(\max(1,\mathrm{substeps}-1)\),
and tool count is the number of distinct action names. These action-derived
proxies support exploratory grouping; the construction is detailed in
Appendix~\ref{app:task_definitions}.

\subsubsection{Airline: Pilot-Scale Evaluation}
\label{subsubsec:design_airline}

Airline provides a second stateful environment involving booking,
itinerary, customer, and payment operations. The evaluated pilot
uses six selected tasks from an environment inventory of fifty
tasks. Its four configurations are Minimal, Planner Fixed,
Planner Sham, and Verifier-only.

The planned design contains
\(5\times6\times4\times2=240\) cells.
The realized dataset contains 233 trajectories, leaving seven
planned runs unobserved. The task identifiers, realized coverage,
and missing cells are reported in
Appendix~\ref{app:airline_results}.

\subsection{Models and Harness Configurations}
\label{subsec:design_models_configurations}

The shared Retail block evaluates the run identifiers
\texttt{claude-haiku},
\texttt{deepseek-v4-flash},
\texttt{deepseek-v4-pro},
\texttt{glm-4-air},
\texttt{glm-4-flash}, and
\texttt{qwen-turbo}.
The planner-focused block evaluates
\texttt{deepseek-v4-flash},
\texttt{doubao-pro},
\texttt{kimi-32k},
\texttt{minimax-text}, and
\texttt{qwen-turbo}.
Airline evaluates
\texttt{claude-haiku},
\texttt{deepseek-v4-flash},
\texttt{doubao-pro},
\texttt{glm-4-air}, and
\texttt{qwen-turbo}.

These are the run keys recorded by the common runtime adapter;
Appendix~\ref{app:model_metadata} lists the available metadata.

Table~\ref{tab:harness_configurations_compact} summarizes the
implemented configurations. There are eight distinct configuration
labels across the study because Planner Self appears only in
the separate planner-focused block.

\begin{table}[t]
\centering
\small
\caption{Harness configurations and experimental availability.}
\label{tab:harness_configurations_compact}
\begin{tabular}{
@{}
P{0.19\textwidth}
P{0.46\textwidth}
P{0.25\textwidth}
@{}
}
\toprule
\textbf{Configuration}
& \textbf{Implemented intervention}
& \textbf{Experimental blocks}
\\
\midrule
Minimal
& Task instruction, tool definitions, and the standard
execution protocol
& All three blocks
\\
Planner Fixed
& Offline task-specific plan containing relevant subgoals
and dependencies
& All three blocks
\\
Planner Sham
& Shuffled policy words matched to the Fixed word count
and supplied in the same wrapper
& All three blocks
\\
Planner Self
& The evaluated model generates a plan before execution
& Retail planner
\\
Evaluator-only
& Evaluation with the configured repair procedure
& Retail shared
\\
Orchestrator-only
& Tool routing, schema, ordering, and execution constraints
& Retail shared
\\
Verifier-only
& Terminal review of proposed completion
& Retail shared; Airline
\\
Full Fixed
& Fixed planning combined with evaluation and repair,
memory, orchestration, and verification
& Retail shared
\\
\bottomrule
\end{tabular}
\end{table}

Full Fixed combines planning, evaluation and repair, memory, orchestration,
and terminal verification. Its comparison with Minimal evaluates this bundled
control stack, including interactions among its components.

Appendix~\ref{app:harness_details} details the configuration semantics,
component timing, and plan text.

\subsection{Fixed Plans, Sham Controls, and Self-Planning}
\label{subsec:design_planning_conditions}

Planner Fixed reads a prewritten plan from a task-indexed file. A given
task uses the same text across models. The implementation requires a
matching plan entry and injects it into system context when the agent
processes the first user message. Planner Sham shuffles words from the
domain policy using seed 2701, then repeats or truncates the sequence to
the Fixed plan's whitespace-word count. Both use a \texttt{<task\_plan>}
wrapper. This is a shuffled word sequence, not a coherent but unhelpful plan.
It controls word count and packaging, not readability or task structure;
tokenizer length can also differ.

Fixed and Sham use static text without a runtime helper-model call.
Plan construction is documented as manual in the authoring script.
The archived text establishes what was injected. The author's full
information access during construction is unrecorded, so interpreting
the result as ordinary planning support assumes that the plans used only
information available before execution. The primary analysis follows an
intention-to-treat principle: episodes remain in their assigned configurations
even if they end before injection.

Some episodes terminate on the user's first reply, before plan injection.
Twenty-five primary pairs contain at least one such episode: 15 Fixed
and 16 Sham rows have empty injection records, all with zero verified
success. Restricting to the 240 pairs with text injected in both arms
gives a 7.50-point difference, close to the primary 7.17 points
(Section~\ref{subsec:planner_robustness}). Appendix~\ref{app:harness_details} gives the exposure counts
and an actual Fixed/Sham example.

Planner Self asks the evaluated model to generate its own plan, adding a
planning call and its token cost. Self-planning and Sham--Minimal contrasts
serve as secondary diagnostics of plan generation and the chosen context control.

\subsection{Recorded Outcomes and Resource Measures}
\label{subsec:design_recorded_measures}

The scoring adapter defines verified success, the completion proxy, and
false pass as in Equation~\ref{eq:observed_outcomes}.
The terminal verifier uses the same model identifier as the executor and
reviews the last at most eight user/assistant messages, truncated to 300
characters each. It receives dialogue text, with no database access, and
has an output limit of 200 tokens. For episodes with \(S=1\), an output
containing \texttt{INCOMPLETE} sets \(R=1\). The episode and oracle reward
are complete before this check; the verifier has no repair loop.

Failure types are rule-based classifications. The adapter assigns
\emph{pseudo-completion} to false passes, \emph{none} to oracle successes,
and otherwise distinguishes budget exhaustion from execution failure
using the termination reason.

Resource records include input/output tokens, tool calls, simulation latency,
component token attribution, and logged monetary cost in USD.
Simulation latency ends before the terminal verifier call.
Cost@Success is mean logged cost divided by the verified-success rate.
Pass@Budget is the conditional verified-success rate among trajectories
whose total tokens are at most the analysis dataset's median \(b\).
For shared Retail, \(b=84{,}388\).

\subsection{Analysis Samples, Matching, and Missingness}
\label{subsec:design_samples}

The primary planning comparison pairs configurations by
model--task--replication key \(c=(m,t,r)\), using cells with both Fixed
and Sham observed. Its estimator is
\begin{equation}
\widehat{\tau}^{P}
=\frac{1}{|\mathcal{C}_{P}|}\sum_{c\in\mathcal{C}_{P}}
\left(Y^{\mathrm{VS}}_{c,\mathrm{fixed}}-Y^{\mathrm{VS}}_{c,\mathrm{sham}}\right).
\label{eq:planner_primary_estimand}
\end{equation}
We first restrict the planner-focused sample to tasks flagged as requiring mutation by the
reference-action write-operation heuristic. This excludes task 62 and
leaves 23 tasks. We then retain models with Minimal verified success
at least 0.05 on this mutation-flagged subset. The rule retains
\texttt{deepseek-v4-flash}, \texttt{doubao-pro}, \texttt{kimi-32k}, and
\texttt{qwen-turbo}, contributing 67, 69, 69, and 60 pairs respectively.
The resulting sample contains 265 matched cells.

\texttt{minimax-text}, whose Minimal success rate is zero, remains in
the five-model results and Holm testing family but falls below the pooled
retention threshold. Its Fixed--Sham contrast has 27 pairs.
The all-task sensitivity retains 277 pairs across the same four models.

Each secondary contrast uses its own available matched cells.
Shared replication identifiers align records; execution order and random
draws were not established as identical across arms. Missing outcomes
remain absent: the planner-focused sample has 213 missing cells relative to its 1,440-cell grid,
and Airline has seven relative to 240 planned cells.

Shared-Retail pooled means describe each configuration's available
trajectories. Model-specific matched Verifier--Minimal false-pass contrasts
are reported in Appendix~\ref{app:verifier_results}. The liability analysis likewise pairs
each configuration to its available Minimal counterparts, then aggregates
within models and risk strata.

\subsection{Statistical Procedures and Evidence Hierarchy}
\label{subsec:design_inference_scope}

The primary pooled Fixed--Sham result uses a 90\% task-clustered
percentile bootstrap interval with 5,000 resamples. Each draw samples
tasks with replacement, retains all matched model--replication cells
within each selected task, and computes their cell-weighted mean.
The pooled and leave-one-model-out sequence starts from seed 2701.
Model-specific primary contrasts use the same nominal level and number
of draws, with seed \(2701+n_{\mathrm{pairs}}\).

Model-specific bootstrap \(p\)-values and the five-model Holm adjustment
are reported together. Some secondary contrasts use matched-cell rather
than task resampling; Appendix~\ref{app:statistical_inference} specifies
the procedure for each analysis.

Model deletion describes sensitivity to the evaluated model composition.
Capability interactions use six measured models and are assessed alongside
the delete-one-model jackknife. The predictive comparison separates
trajectory-derived cross-fitted proxies from the independent held-out
probe sensitivity. Both are model-level generalization checks.

The economic grid assigns \(V\in\{1,5,20\}\) and \(L\in\{2,200\}\),
in USD. Within each task-risk stratum, scenario summaries give equal
weight to contributing models and the three \(V\) values at each \(L\).
Risk is a reference-action heuristic defined in
Section~\ref{subsec:liability_scenario_inputs}; the low-risk stratum
contains one task. Break-even values use the aggregate differences.

The experimental package records a 7 July 2026 freeze
and a 13 July 2026 coverage audit.

\section{Planning Information: Evidence from the Fixed--Sham Comparison}
\label{sec:planner_results}

The planner-focused Retail experiment provides the primary Fixed--Sham
comparison. We report the pooled result, secondary contrasts, heterogeneity,
and sensitivity, followed by supplementary shared-Retail and Airline results.

\subsection{Average Planning Effect in the Retained Matched Sample}
\label{subsec:planner_average}

\begin{figure}[t]
\centering
\includegraphics[width=0.92\linewidth]{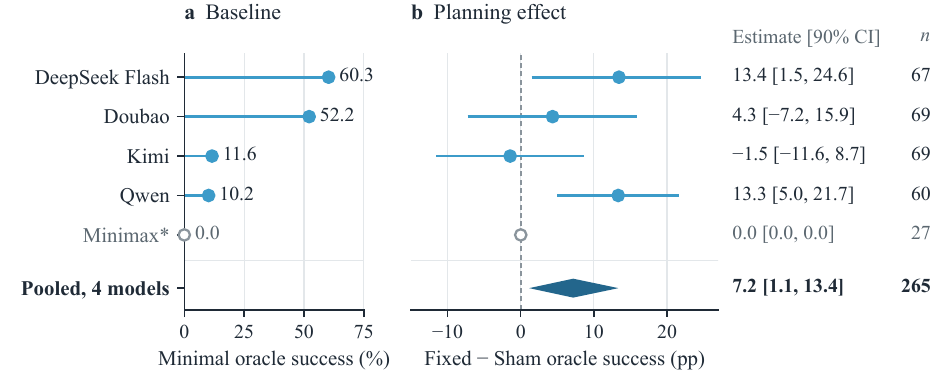}
\caption{\textbf{Baseline success and Fixed--Sham planning effects in Retail (E03A).}
(a) Minimal oracle-success rate by model; rows in both panels are ordered by
this baseline. (b) Model-specific Fixed--Sham differences in oracle success
with 90\% task-clustered percentile bootstrap intervals (5,000 draws); the
diamond shows the pooled result for four models and 265 matched cells across
23 tasks. The right-hand columns give each estimate and interval in
percentage points and the number of matched cells. Qwen and Kimi have similar
baselines but different planning responses. *Minimax is outside this pool:
both matched arms have zero successes, producing a degenerate interval at
zero. Model-specific Holm-adjusted quantities appear in
Table~\ref{tab:planner_model_effects_main}.}
\label{fig:planner_fixed_sham}
\end{figure}

The primary comparison averages 265 matched cells from four models and
23 mutation-flagged tasks using Equation~\ref{eq:planner_primary_estimand}:
\begin{equation}
\widehat{\tau}^{P}=0.0717,\qquad
\mathcal I_{\mathrm{task},90\%}=[0.0115,\,0.1336].
\label{eq:planner_average_result}
\end{equation}
Fixed raises oracle-verified success by 7.17 percentage points relative
to Sham. The 90\% task-clustered percentile interval
is positive. Figure~\ref{fig:planner_fixed_sham} presents the pooled
and model-specific estimates.

\subsection{Sham--Minimal Placebo and Secondary Planning Contrasts}
\label{subsec:planner_secondary}

The Sham--Minimal comparison examines the behavior of
the chosen context control. Across 264 matched cells,
the pooled result is
\begin{equation}
\widehat{\tau}^{\mathrm{sham-min}}
=
0.0189,
\qquad
\mathcal{I}_{\mathrm{task}}^{\mathrm{sham-min}}
=
[-0.0345,\,0.0709].
\label{eq:planner_placebo_result}
\end{equation}
Sham has a positive point estimate relative to Minimal, providing no
average decrement from the shuffled control. Its interval spans
zero, so this diagnostic does not establish equivalence.
Table~\ref{tab:app_planner_placebo} reports the model-specific results.

Fixed--Minimal estimates are positive for all four retained models,
ranging from 5.80 to 11.59 percentage points. Thus, the favorable Fixed
point estimates also appear against the no-added-plan baseline. Self-planning varies by executor: Qwen
improves over Sham by 16.95 pp (interval [6.78, 27.12], unadjusted
\(p=0.0052\), 59 pairs), while its direct Self--Fixed estimate is 3.39 pp
with interval [-8.47, 15.25]. The other retained models have negative
Self--Fixed point estimates and intervals that include or touch zero.
Self also adds a planning call. Full contrasts and effort records are in
Appendix~\ref{app:planner_results}.

Fixed--Sham false-pass estimates are negative for all four retained models,
ranging from -11.94 to -4.35 pp, with most intervals touching or crossing
zero (Table~\ref{tab:app_planner_fp}). Planning guidance thus affects
both oracle success and erroneous completion.

\subsection{Heterogeneity across Models and Task Complexity}
\label{subsec:planner_heterogeneity}

Table~\ref{tab:planner_model_effects_main} shows variation in the magnitude
and direction of planner-focused Retail Fixed--Sham effects.

\begin{table}[t]
\centering
\small
\caption{
Model-specific Fixed--Sham effects on verified
success in planner-focused Retail. The pooled estimate uses the four
models other than \texttt{minimax-text}.
Bounds are 90\% task-clustered bootstrap intervals; Holm adjustment
uses all five model-specific contrasts. Effects are on the proportion
scale; multiply by 100 for percentage points.
}
\label{tab:planner_model_effects_main}
\setlength{\tabcolsep}{4pt}
\begin{tabular}{@{}lrrrrrr@{}}
\toprule
\textbf{Model}
& \textbf{\(n\)}
& \textbf{Effect}
& \textbf{Lower}
& \textbf{Upper}
& \textbf{\(p\)}
& \textbf{Holm \(p\)}
\\
\midrule
\texttt{deepseek-v4-flash}
& 67 & 0.1343 & 0.0149 & 0.2464 & 0.08 & 0.32
\\
\texttt{doubao-pro}
& 69 & 0.0435 & -0.0725 & 0.1594 & 0.5992 & 1.0
\\
\texttt{kimi-32k}
& 69 & -0.0145 & -0.1159 & 0.087 & 0.9288 & 1.0
\\
\texttt{minimax-text}
& 27 & 0.0 & 0.0 & 0.0 & 1.0 & 1.0
\\
\texttt{qwen-turbo}
& 60 & 0.1333 & 0.05 & 0.2167 & 0.0172 & 0.086
\\
\bottomrule
\end{tabular}
\end{table}

DeepSeek Flash and Qwen have the largest positive point estimates.
Qwen's unadjusted \(p\)-value is 0.0172 and Holm-adjusted value is
0.086. All five adjusted values exceed 0.05.

Similar baseline success can accompany different planning responses.
Qwen has Minimal success of 10.17\% and a Fixed--Sham effect of +13.33 pp;
Kimi has Minimal success of 11.59\% and an effect of -1.45 pp
(Figure~\ref{fig:planner_fixed_sham}).

The task-complexity analysis shows a related pattern.
Median splits on reference-action count, the dependency-depth
proxy, and distinct tool-name count produce the
same partition of the retained sample:
\begin{equation}
\widehat{\tau}^{P}_{\mathrm{high}}
=
0.1301,
\qquad
\widehat{\tau}^{P}_{\mathrm{low}}
=
0.0211.
\label{eq:planner_complexity_result}
\end{equation}
The high-complexity group contains 123 matched
cells and the low-complexity group contains 142
(Table~\ref{tab:app_planner_complexity}).
The recorded median cutoffs are 4 for substeps,
3 for dependency depth, and 3 for tool count.
The corresponding continuous slopes
are \(0.0178\), \(0.0172\), and \(0.0318\).

The higher-complexity group has the larger contrast. The three
proxies define one coincident partition of 23 tasks, so this is a single
exploratory grouping result.

\subsection{Sensitivity to Model Composition and Task Inclusion}
\label{subsec:planner_robustness}

Table~\ref{tab:planner_lomo_main} reports the pooled estimate and its
sensitivity to omitting each retained model.

\begin{table}[t]
\centering
\small
\caption{
Pooled and leave-one-model-out Fixed--Sham estimates
with 90\% task-clustered bootstrap intervals. Every row retains
twenty-three tasks. Model deletions refer to the
four-model pooled analysis sample. Effects are on the proportion
scale; multiply by 100 for percentage points.
}
\label{tab:planner_lomo_main}
\setlength{\tabcolsep}{5pt}
\begin{tabular}{@{}lrrrr@{}}
\toprule
\textbf{Analysis scope}
& \textbf{Cells}
& \textbf{Effect}
& \textbf{Lower}
& \textbf{Upper}
\\
\midrule
All retained models
& 265 & 0.0717 & 0.0115 & 0.1336
\\
Drop \texttt{deepseek-v4-flash}
& 198 & 0.0505 & -0.0202 & 0.1179
\\
Drop \texttt{doubao-pro}
& 196 & 0.0816 & 0.0158 & 0.1515
\\
Drop \texttt{kimi-32k}
& 196 & 0.1020 & 0.0402 & 0.1667
\\
Drop \texttt{qwen-turbo}
& 205 & 0.0537 & -0.0149 & 0.1256
\\
\bottomrule
\end{tabular}
\end{table}

Every deletion-specific point estimate remains
positive, ranging from 5.05 to 10.20 pp.
The interval crosses zero after omitting
either \texttt{deepseek-v4-flash} or
\texttt{qwen-turbo}. The direction is
therefore stable across these deletions, while
the interval-based evidence depends on model
composition.

The all-task sensitivity gives 7.58 pp across 277 pairs. Restricting to
the 240 pairs with nonempty injections in both arms gives 7.50 pp
(90\% task-clustered interval, [0.91, 14.23]), a change of just
0.33 pp from the primary estimate. This exposure-restricted
sensitivity retains all 23 tasks. Replication-specific estimates are
5.62, 6.82, and 9.09 pp. Appendix~\ref{app:planner_results} gives
the complete sensitivity results.

\subsection{Supplementary Evidence from Shared Retail and Airline}
\label{subsec:planner_supplementary}

The supplementary shared-Retail analysis evaluates Fixed--Sham across all six models. Five
point estimates are positive, ranging from 6.67 to 20.00 pp; GLM Flash
has an estimate of -2.22 pp. GLM Air has the strongest unadjusted result,
20.00 pp with a 90\% interval of [6.67, 33.33] and \(p=0.012\).
This comparison uses a different task and model sample from the primary planner analysis
(Table~\ref{tab:app_planner_e03b}).

In Airline, DeepSeek Flash has the largest Fixed--Sham estimate, 25.00 pp,
with a 90\% interval of [8.33, 41.67] across 12 pairs. Its unadjusted \(p\)-value
is 0.028 and Holm-adjusted value is 0.14. None of the five Airline
Fixed--Sham comparisons passes the 0.05 adjusted threshold.
Appendix~\ref{app:airline_results} reports the complete pilot results,
including the cross-environment comparison in
Figure~\ref{fig:app_cross_domain}.

\section{Verification and False-Pass Outcomes}
\label{sec:verifier_results}

The terminal verifier changes which completed episodes are accepted.
We first report what it rejects, then compare matched configurations
and examine model heterogeneity and the Airline pilot.

\subsection{Rejection Outcomes and Retail Comparisons}
\label{subsec:verifier_retail_descriptive}

The verifier rejects 83 of 137 oracle-invalid Retail episodes (61\%)
and 16 of 92 oracle-correct episodes (17\%). Of its 99 rejections,
83 concern invalid outcomes (84\%). Table~\ref{tab:app_verifier_outcome_decomposition}
shows the counts, including Full Fixed and Airline.

\begin{table}[!htbp]
\centering\small
\caption{Post-hoc rejection by oracle validity. Each cell is a trajectory count.}
\label{tab:app_verifier_outcome_decomposition}
\begin{tabular}{@{}llrrrrr@{}}
\toprule Domain & Configuration & Total & \(Z=1\) & \(Z=0\) & \(R=1,Z=1\) & \(R=1,Z=0\)\\
\midrule
Retail & Verifier-only & 229 & 92 & 137 & 16 & 83\\
Retail & Full Fixed & 142 & 51 & 91 & 5 & 38\\
Airline & Verifier-only & 58 & 17 & 41 & 7 & 25\\
\bottomrule
\end{tabular}
\end{table}

Of the 137 oracle-invalid episodes, 83 are rejected and 54 are not.
Six of those 54 have \(S=0\) (the agent did not assert completion),
leaving 48 false passes. Setting \(R=0\) while holding these executions
fixed recovers the pre-check erroneous-completion rate:
\((48+83)/229=57.21\%\). The actual false-pass rate is
\(48/229=20.96\%\), so rejection removes 36.24 percentage points.
Minimal has a false-pass rate of 58.30\%. The cross-arm gap decomposes as
\begin{equation}
\underbrace{P^{\mathrm{FP}}_{\mathrm{Minimal}}-P^{\mathrm{FP}}_{\mathrm{Verifier}}}_{37.34\ \mathrm{pp}}
=\underbrace{P^{\mathrm{FP}}_{\mathrm{Minimal}}-P^{\mathrm{FP}}_{\mathrm{Verifier},R=0}}_{1.09\ \mathrm{pp}}
+\underbrace{P^{\mathrm{FP}}_{\mathrm{Verifier},R=0}-P^{\mathrm{FP}}_{\mathrm{Verifier}}}_{36.24\ \mathrm{pp}}.
\label{eq:verifier_false_pass_main}
\end{equation}
Almost all of the pooled gap is
accounted for by the rejection flags.

The six matched Retail Verifier--Minimal false-pass estimates are all
negative (Table~\ref{tab:app_retail_verifier_matched}). Reductions range
from 6.25 points for DeepSeek Pro to 100 points for Claude; Claude's
estimate uses ten pairs. The pooled Minimal and Verifier-only oracle-success rates are
37.87\% and 40.17\%; Table~\ref{tab:verifier_retail_main} gives the
full comparison.

\begin{figure}[!htbp]
\centering
\includegraphics[width=0.92\linewidth]{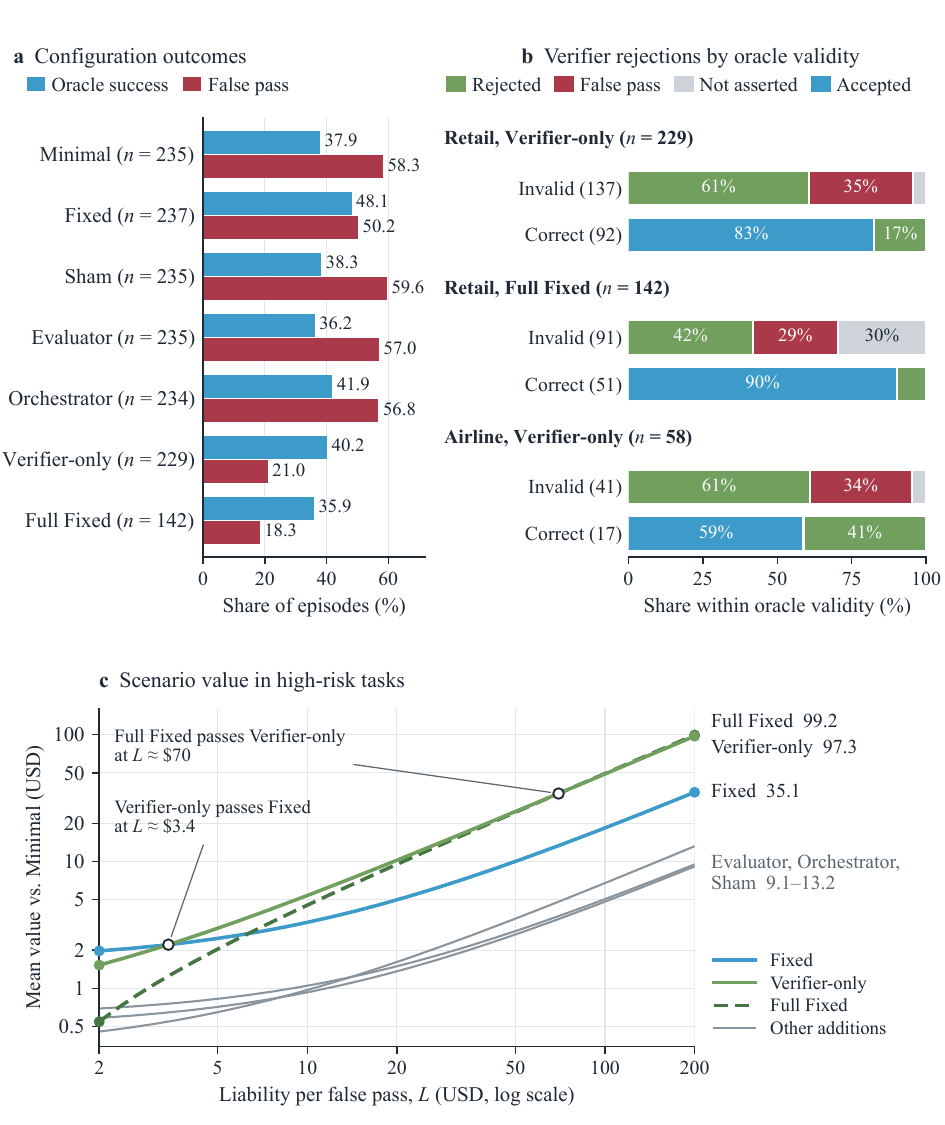}
\caption{\textbf{Configuration outcomes, verifier rejections, and liability scenarios.}
(a) Pooled oracle-success and false-pass rates for all seven shared-Retail
configurations; \(n\) counts trajectories, and coverage varies by
configuration. (b) Verifier outcomes within oracle-invalid (\(Z=0\)) and
oracle-correct (\(Z=1\)) episodes: rejected (\(R=1\)), false pass, not
asserted (\(S=0\)), and accepted. Row labels give episode counts; the counts
are in Table~\ref{tab:app_verifier_outcome_decomposition}. (c) Mean scenario
value of each addition relative to Minimal in the high-risk stratum as a
function of the liability per false pass \(L\), on log scales. Markers at
\(L=2\) and \(L=200\) are the frozen grid values; because scenario value is
linear in \(L\), the curves between them follow directly. Values weight the
six models and \(V\in\{1,5,20\}\) equally. The marked crossing points compare
scenario functions estimated on configuration-specific matched samples and are
descriptive. Pair counts and the full risk grid are in
Tables~\ref{tab:app_e05_pair_coverage} and~\ref{tab:app_liability_full}.}
\label{fig:verifier_economics}
\end{figure}

Figure~\ref{fig:verifier_economics} brings together configuration outcomes,
the verifier's rejections by oracle validity, and the liability scenarios in
Section~\ref{sec:liability_analysis}.

\subsection{Rule-Based Failure Labels and Exploratory Model Heterogeneity}
\label{subsec:verifier_failure_heterogeneity}

The failure taxonomy summarizes the scoring outcomes of the planning arms
(Table~\ref{tab:app_planner_failure_types}). Pseudo-completion is the
largest category in each arm and is assigned directly from false pass;
the remaining labels partition oracle success, budget exhaustion, and
execution failure.

We also examine how the independent Failure Detection Sensitivity (FDS) probe score relates to verifier
exposure in the 1,547 shared-Retail trajectories. The fitted interaction is
\begin{equation}
\widehat\beta_{\mathrm{FDS}\times V}=0.1825.
\label{eq:fds_verifier_main}
\end{equation}
With false pass as the outcome, the positive coefficient associates higher
measured FDS with a smaller additional reduction under verification.
Across six model deletions the coefficient remains positive, ranging from
0.1492 to 0.5078; its largest value occurs after dropping GLM Flash.
The delete-one-model jackknife gives SE 0.2803 and \(p=0.5439\),
indicating substantial uncertainty in this six-model association.
Full coefficients and diagnostics appear in
Tables~\ref{tab:app_verifier_interaction} and~\ref{tab:app_verifier_lomo}.

\subsection{Airline Pilot Evidence}
\label{subsec:verifier_airline}

In Airline, the verifier rejects 25 of 41 oracle-invalid episodes (61\%)
and seven of 17 oracle-correct episodes (41\%). The correct-episode
rejection fraction is higher than Retail's 17\%
(Table~\ref{tab:app_verifier_outcome_decomposition}).

The Airline matched Verifier--Minimal contrasts show large false-pass
reductions for \texttt{claude-haiku} (-70.00 pp) and \texttt{qwen-turbo}
(-91.67 pp). DeepSeek Flash and GLM Air each have a -25.00 pp estimate,
with intervals crossing zero. Doubao's estimate is +10.00 pp, also
with an interval crossing zero. Paired counts range from ten to twelve.

Table~\ref{tab:app_airline_verifier} gives the 90\% task-clustered intervals
and bootstrap quantities. Figure~\ref{fig:app_cross_domain} in
Appendix~\ref{app:airline_results} compares the model-specific patterns
across environments.

\subsection{What the Terminal Verifier Changes}
\label{subsec:verifier_interpretation_limits}

In Retail, the terminal verifier removes 83 of 137 oracle-invalid episodes
(61\%) and withholds 16 of 92 oracle-correct ones (17\%), reducing the
false-pass rate from 57.21\% to 20.96\%. The additional inference cost
is less than one cent per episode (\(\overline{\Delta C}=0.0079\) USD
for Verifier-only). The direct counts in
Table~\ref{tab:app_verifier_outcome_decomposition} give the full
cross-tabulation.

\section{Configuration Costs and Liability Scenarios}
\label{sec:liability_analysis}

Shared-Retail outcomes reveal different trade-offs between oracle success,
false pass, and resource use. The liability-scenario analysis values
paired configuration differences under a specified USD scenario grid.

\subsection{Descriptive Performance and Resource Trade-offs}
\label{subsec:configuration_resource_tradeoffs}

Table~\ref{tab:configuration_economics_main} reports all seven pooled
configuration rows from the shared Retail experiment.

\begin{table*}[t]
    \centering
    \small
    \caption{
    Pooled Retail configuration outcomes and economic measures.
    Samples are unbalanced; comparisons are not paired.
    Cost@Success is cost per oracle success; Pass@Budget conditions on total
    tokens at or below 84,388.
    }
    \label{tab:configuration_economics_main}
    \setlength{\tabcolsep}{4pt}
    \begin{adjustbox}{max width=\linewidth}
    \begin{tabular}{@{}lrrrrrr@{}}
        \toprule
        \textbf{Configuration}
        & \textbf{\(n\)}
        & \textbf{Verified}
        & \textbf{False pass}
        & \textbf{Mean cost}
        & \textbf{Cost@Success}
        & \textbf{Pass@Budget} \\
        \midrule
\texttt{minimal}
        & 235 & 0.3787 & 0.5830 & 0.0136 & 0.0360 & 0.4722 \\
\texttt{planner\_fixed}
        & 237 & 0.4810 & 0.5021 & 0.0135 & 0.0280 & 0.5185 \\
\texttt{planner\_sham}
        & 235 & 0.3830 & 0.5957 & 0.0144 & 0.0377 & 0.4697 \\
\texttt{evaluator\_only}
        & 235 & 0.3617 & 0.5702 & 0.0268 & 0.0741 & 0.3735 \\
\texttt{orchestrator\_only}
        & 234 & 0.4188 & 0.5684 & 0.0147 & 0.0350 & 0.4569 \\
\texttt{verifier\_only}
        & 229 & 0.4017 & 0.2096 & 0.0164 & 0.0409 & 0.5086 \\
\texttt{full\_fixed}
        & 142 & 0.3592 & 0.1831 & 0.0739 & 0.2058 & 0.3333 \\
\bottomrule
    \end{tabular}
    \end{adjustbox}
\end{table*}

Within these pooled summaries, Planner Fixed has the highest verified
success (\(0.4810\)), the lowest Cost@Success (\(0.0280\)),
and the highest Pass@Budget (\(0.5185\)).
Verifier-only has a lower false-pass rate (\(0.2096\)), with verified
success of \(0.4017\) and mean logged cost of \(0.0164\).
Full Fixed has the lowest pooled false-pass rate (\(0.1831\)), but its
verified success is \(0.3592\) and its mean logged cost is \(0.0739\),
the largest configuration mean.

These pooled means use 142 trajectories for Full Fixed, 235 for Minimal,
and 237 for Planner Fixed. Their different coverage motivates the paired
comparisons used in the scenario analysis.

The model-level resource records also show that execution requirements
vary across configurations. For \texttt{deepseek-v4-flash}, the
mean token counts are \(73{,}409.7\) under Minimal, \(75{,}102.9\) under
verifier-only, and \(250{,}745.1\) under Full Fixed. The complete model-level
outcomes, token counts, and component token shares remain in
Appendix~\ref{app:configuration_economics}.

Pass@Budget conditions on total tokens at or below the shared-Retail
median of 84,388.

\subsection{Scenario Inputs and Incremental Net Value}
\label{subsec:liability_scenario_inputs}

The liability analysis values configuration--Minimal differences using
Equation~\ref{eq:component_net_value}. Within each model and risk stratum,
we pair observed task--replication cells and calculate the mean success
difference, avoided false pass, and incremental logged cost. These
model-level differences are expanded over
\begin{equation}
V_t\in\{1,5,20\},\qquad L_t\in\{2,200\},
\label{eq:scenario_grid}
\end{equation}
in USD. For each configuration and \(L_t\), the summary averages
equally over contributing models and the three \(V_t\) values.
Thus \(n=18,15,9\) counts scenario rows from six, five, or three models,
respectively. Appendix~\ref{app:liability_results} gives the matched-pair
coverage and the full scenario table.

The high-risk flag is assigned when a reference action name contains
\texttt{cancel}, \texttt{modify}, \texttt{exchange}, \texttt{return},
\texttt{refund}, \texttt{update}, \texttt{place}, or \texttt{delete};
otherwise the task is labeled low risk. This action-based grouping is
separate from the imposed liability weight \(L_t\). The low-risk group
contains only task 62, also flagged as read-only by the mutation heuristic.
We focus the main comparison on the high-risk stratum and report the
single-task sensitivity in the appendix.

The scenario payoff assigns \(+V_t\) to oracle success and \(-L_t\)
to false pass.

\subsection{Reported Configuration Values under Alternative Scenarios}
\label{subsec:liability_rankings}

Table~\ref{tab:liability_selected} compares three configurations in the
high-risk task stratum at the two liability weights. The complete six-configuration,
two-risk grid and component differences appear in Appendix~\ref{app:liability_results}.

\begin{table}[t]\centering\small
\caption{High-risk E05 scenario values relative to Minimal.
Means weight six contributing models and three values of \(V\) equally.
Matched pairs vary by configuration; the full grid is in
Table~\ref{tab:app_liability_full}.}
\label{tab:liability_selected}
\begin{tabular}{@{}lrrr@{}}
\toprule Configuration & \(L\) (USD) & Mean NV (USD) & Matched pairs\\\midrule
Planner Fixed & 2 & 1.9748 & 217 \\
Planner Fixed & 200 & 35.1299 & 217 \\
Verifier-only & 2 & 1.5275 & 214 \\
Verifier-only & 200 & 97.3133 & 214 \\
Full Fixed & 2 & 0.5460 & 135 \\
Full Fixed & 200 & 99.1962 & 135 \\
\bottomrule\end{tabular}\end{table}

The ordering changes between the two liability levels. At \(L=2\),
Fixed has the largest mean value, followed by Verifier-only and Full Fixed.
At \(L=200\), Full Fixed and Verifier-only have the largest values.
Figure~\ref{fig:verifier_economics}(c) extends this comparison to all
six non-Minimal configurations and to liability values between the two
grid points.

The source of this change is visible in the component
differences. For Full Fixed in the high-risk stratum,
\begin{equation}
    \overline{\Delta P}
    =
    -0.0411,
    \qquad
    \overline{\Delta FP}^{\mathrm{avoid}}
    =
    0.4982,
    \qquad
    \overline{\Delta C}
    =
    0.0947.
    \label{eq:full_fixed_components}
\end{equation}
The mean verified-success difference is negative and incremental cost
is positive. Its positive scenario value therefore relies on the
weight assigned to the avoided false-pass term.

For Verifier-only in the high-risk stratum, the components are
\begin{equation}
    \overline{\Delta P}
    =
    0.0655,
    \qquad
    \overline{\Delta FP}^{\mathrm{avoid}}
    =
    0.4838,
    \qquad
    \overline{\Delta C}
    =
    0.0079.
    \label{eq:verifier_components}
\end{equation}
Verifier-only and Full Fixed avoid nearly the same false-pass share
(48.38 vs.\ 49.82 pp), but Full Fixed has negative \(\overline{\Delta P}\)
and twelve times the incremental cost. The standalone verifier therefore
captures nearly all the avoided false-pass benefit of the full stack at
a fraction of its cost.

The terminal verifier acts after oracle scoring, so \(\overline{\Delta P}\)
reflects the arms' executions rather than a gain produced by the check.
Setting Verifier-only's success term to zero lowers its value to \$0.96
at \(L=2\) and \$96.75 at \(L=200\). The ordering in
Table~\ref{tab:liability_selected}, and across all six configurations,
remains unchanged.

\subsection{Break-Even Liability Relative to Minimal}
\label{subsec:break_even_liability}

Equation~\ref{eq:framework_break_even} gives the liability at which a
configuration's aggregate scenario value reaches zero relative to Minimal.
We apply it to the model-equal mean differences in each risk stratum.

Full Fixed in the high-risk stratum has thresholds 0.2724, 0.6020, and
1.8378 USD at \(V=1,5,20\). Its success difference is negative, so a
higher success value requires more avoided false-pass liability to offset
that difference and the additional inference cost. Planner Fixed and
Verifier-only have negative thresholds at all three values, corresponding
to positive scenario value at \(L=0\).

Evaluator-only also has a small positive threshold, 0.0021 USD, at
\(V=1\) in the high-risk stratum. Table~\ref{tab:app_break_even_full}
reports every threshold. Each is a comparison with Minimal on its own
paired sample; pairwise configuration crossover values would use
differences between two configurations' scenario functions.

\subsection{Sample Dependence and Economic Interpretation}
\label{subsec:liability_sample_interpretation}

The pooled rates in Table~\ref{tab:configuration_economics_main} weight
each observed trajectory equally. The liability analysis first computes paired differences
within models and task-risk strata, then weights contributing models
equally. These different samples and weights explain why Fixed has a
lower pooled mean cost than Minimal while its high-risk paired
incremental cost is positive. 

All added configurations cover their incremental inference cost at the
evaluated liability levels: every aggregate break-even threshold lies below
the grid's lower bound of \(L=2\). The informative comparison is therefore
how configurations rank as liability changes, rather than the positivity
of all 24 grid values.

Logged execution cost covers the recorded inference expenditure.
Plan authoring, integration, maintenance, human review, and the
opportunity cost of withholding correct outcomes require additional
accounting. These quantities must be measured for a deployment-level cost comparison.

\section{Discussion}
\label{sec:discussion}

The results connect the information supplied to an executor, the acceptance
of its completed work, and the value assigned to those outcomes. These
connections suggest concrete changes to component evaluation and identify
where further evidence is needed.

\subsection{Implications for Component-Level Agent Evaluation}
\label{subsec:benchmark_implications}

Component evaluation benefits from controls tailored to the intervention.
Fixed--Sham holds word count and packaging comparable while changing the
supplied guidance. Its positive pooled result motivates testing the
usefulness of task-specific content separately from the cost and quality
of generating a plan. The heterogeneous Self contrasts reinforce the need
to evaluate plan provision and plan generation as distinct procedures.

Verification requires direct measurement of correctness and acceptance.
The tested check withholds invalid episodes and some correct episodes;
oracle success alone misses this trade-off. Reporting their joint counts,
alongside tokens and cost, makes the operating consequences visible.

\subsection{Implications for Planning and Verification Design}
\label{subsec:production_design_implications}

Useful guidance must be available at the time of execution and usable by
the executor. The variation across models and reference-action
complexity groups motivates prospective tests on broader, balanced samples.
Such tests can assess whether the planning patterns predict gains
on new tasks and whether generating comparable guidance justifies its cost.

Terminal verification regulates reliance on completed work. Following
rejected cases through repair, escalation, or abandonment would measure
both the benefit of withholding invalid results and the cost of withholding
valid ones.

Model-level patterns warrant targeted follow-up.
Kimi and Qwen share similar baselines yet respond oppositely to
planning (Section~\ref{subsec:planner_heterogeneity}), and individual
models can dominate specific contrasts. These cases illustrate why
model-specific patterns, not only pooled averages, are needed to judge
harness value.

The liability scenarios make this design dependence explicit: greater
weight on false passes raises the relative value of configurations that
avoid them.

\subsection{Capability Measurement and Harness Selection}
\label{subsec:capability_measurement_selection}
\label{sec:measurement_audit}

The capability audit finds strong Planning Frontier (PF) split-half
reliability (0.968), uncertain Failure Detection Sensitivity (FDS) reliability
(0.761; split range [-0.739,0.931]), and weak Tool Coordination Complexity
(TCC) measurement. TCC reliability rises from -2.116 to 0.349 after adding
stress items, with continued score saturation. The PF strict audit supports
relative comparisons while revealing systematically higher primary scores
(Appendices~\ref{app:capability_probes} and~\ref{app:tcc_failure}).

Neither feature source establishes a reliable predictive advantage in
this six-model comparison. Trajectory-derived proxies worsen all three
metrics. Independent probes change \(R^2\) by +0.002, AUC by +0.010,
and Brier by -0.0005, leaving performance close to baseline
(Table~\ref{tab:app_incremental_validity}). The proxy features also use each held-out model's
Minimal outcomes from other task folds, so they require calibration
trajectories from that model. Harness selection would instead require
predicting configuration benefits and costs.

\subsection{Limitations and Generalizability}
\label{subsec:scope_conditions}

The primary sample is selected by Minimal success and task mutation status;
missing cells limit coverage. Airline contains six tasks, and both
environments use public benchmarks, so participating models may have
encountered these tasks during training. Broader tests on held-out
tasks are needed to assess transfer.

The exposure-restricted
estimate is close to the primary result.

Exact hosted endpoint revisions and some request settings are missing,
and analysis choices lack an independently timestamped preregistration.
The available runtime metadata are listed in Appendix~\ref{app:model_metadata}.

\subsection{Future Research}
\label{subsec:future_research}

Balanced component comparisons with documented plan provenance and
follow-up of rejected cases would test these patterns prospectively.
Broader coverage and more discriminating tool-coordination probes would
support prediction of configuration benefits on new models and tasks.

\section{Conclusion}
\label{sec:conclusion}

Prewritten task-specific guidance improves oracle-verified success over
the matched Sham control by 7.17 percentage points
(90\% interval, 1.15--13.36 points), with gains concentrated in
higher-complexity tasks and without an additional runtime planning call.
A read-only terminal verifier rejects 61\% of Retail oracle-invalid episodes
while withholding 17\% of correct ones, at less than one cent per episode.
The standalone verifier captures nearly all the avoided false-pass benefit
of the full planning-plus-verification stack at a twelfth of its
incremental cost; which component delivers more value depends on the
loss assigned to erroneous acceptance.

\begingroup
\emergencystretch=1em
\bibliographystyle{aer}
\bibliography{paper3}
\endgroup

\clearpage
\appendix

\section{Experimental Protocol and Evidence Hierarchy}
\label{app:protocol}

\subsection{Protocol Record and Scope}

The experimental package was frozen on 7 July 2026, followed by a coverage
audit dated 13 July 2026. The appendices document the protocol, complete
results, sensitivity analyses, and data-quality checks. Analyses that reuse
the same trajectories are identified in Table~\ref{tab:app_protocol_map}.

\subsection{Experiment Map}

Table~\ref{tab:app_protocol_map} distinguishes data collection from analyses
that reuse those data. E03B and the trajectory components of E04--E06 use
the shared Retail block. The capability probes and scalar
benchmark measurements are separate inputs to the relevant analyses.

\begin{table*}[!htbp]
\centering
\small
\setlength{\tabcolsep}{4pt}
\caption{Experiment map and analysis roles. Replication counts describe the design; realized coverage is incomplete in some blocks.}
\label{tab:app_protocol_map}

\begin{adjustbox}{max width=\linewidth,max totalheight=0.82\textheight}
\begin{tabular}{@{}P{0.055\textwidth}P{0.17\textwidth}P{0.30\textwidth}P{0.20\textwidth}P{0.17\textwidth}@{}}
\toprule
\textbf{ID} & \textbf{Object} & \textbf{Realized coverage} & \textbf{Primary output} & \textbf{Status}\\
\midrule
E01 & \(PF/TCC/FDS\) probes & 486 rows; 6 models; 3 axes; 9 items; 3 reps & reliability and PF audit & PF supported; FDS uncertain; TCC failed\\
E02 & Retail configuration economics & 1,547 rows; 6 models; 16 tasks; 7 configs; 3 reps & success, false pass, cost & descriptive configuration panel\\
E03A & Retail planner intervention & 1,227 rows; 5 models; 24 tasks; 4 arms; 3 reps & fixed--sham effect & primary planner analysis\\
E03B & Shared-Retail planner contrast & subset of E02 shared data; 6 models & model-specific fixed--sham & supplementary heterogeneity\\
E04 & \(FDS\times\)Verifier & 1,547 trajectories; 6 model clusters & false-pass interaction & exploratory, few-model inference\\
E05 & Liability scenarios & matched Retail cells; \(V_t\in\{1,5,20\}\), \(L_t\in\{2,200\}\) & paired net value & scenario analysis\\
E06 & Incremental validity & 6 models; 72 Bench observations per model & LOMO \(R^2\), AUC, Brier & feature-source comparison\\
E07 & Airline external validity & 233 rows; 5 models; 6 tasks; 4 configs; 2 reps & planner and verifier contrasts & pilot-scale\\
E08 & TCC stress probe & 198 attempts; 171 valid; 11 models; 6 items; 3 reps & merged reliability & measurement failure\\
\bottomrule
\end{tabular}
\end{adjustbox}
\end{table*}

\subsection{Evidence Hierarchy}

The primary planning comparison is Fixed--Sham in the E03A sample defined
in Section~\ref{subsec:design_samples}. E03B is supplementary shared-Retail evidence. Model-level
comparisons, complexity patterns, and model-composition sensitivity are
reported separately from the pooled planning average.

Verification is assessed through rejection counts, matched contrasts,
and pooled configuration rates. E04 is
exploratory across six model clusters. E05 is a conditional scenario
analysis. E06 compares trajectory-proxy and held-out-probe predictive
inputs, E07 provides second-environment pilot evidence, and E08 evaluates
tool-coordination measurement.

Primary and secondary designate analysis roles. Infrastructure invocation
failures are recorded separately from behavioral task outcomes.

\FloatBarrier

\section{Task Definitions, Environments, and Oracles}
\label{app:task_definitions}

\subsection{Environments and Reported Task Inventory}

Retail covers customer-service tasks involving order changes, returns,
exchanges, payment and account operations, and fulfillment. Airline covers
booking, itinerary, customer, and payment workflows.

Table~\ref{tab:app_task_inventory} gives task counts and the six evaluated
Airline task identifiers. The Airline environment inventory contains
fifty tasks.

\begin{table*}[!htbp]
\centering
\small
\caption{Task inventory in the frozen experiments.}
\label{tab:app_task_inventory}

\begin{adjustbox}{max width=\linewidth,max totalheight=0.82\textheight}
\begin{tabular}{llr}
\toprule
\textbf{Domain} & \textbf{Experimental block} & \textbf{Tasks used}\\
\midrule
Retail & shared seven-configuration block & 16\\
Retail & planner four-arm block & 24\\
Airline & released environment inventory & 50\\
Airline & realized pilot (IDs 8, 14, 16, 18, 20, 21) & 6\\
\bottomrule
\end{tabular}
\end{adjustbox}
\end{table*}

\subsection{Environment State and Oracle Assessment}

The adapter runs the Retail and Airline environments from \(\tau^2\)-bench.
It records the benchmark reward and defines oracle success as
\(Z=1[\mathrm{reward}\geq0.999]\). Termination supplies the completion
proxy \(S\), and the dialogue-based terminal verifier supplies rejection
\(R\). Equation~\ref{eq:observed_outcomes} gives the scoring relationship.
The joint rejection counts in Appendix~\ref{app:verifier_results} follow
directly from these row-level fields.

\subsection{Task-Complexity Variables}

The adapter constructs three descriptors from each task's reference actions:
\begin{equation}
N_t^{\mathrm{step}}=|\mathrm{actions}_t|,\quad
D_t^{\mathrm{dep}}=\max(1,N_t^{\mathrm{step}}-1),\quad
N_t^{\mathrm{tool}}=|\mathrm{unique\ action\ names}_t|.
\label{eq:app_task_complexity}
\end{equation}
Dependency depth is therefore an action-count proxy.
Table~\ref{tab:app_planner_complexity} reports the
estimates and continuous slopes.

\section{Models and Available Runtime Metadata}
\label{app:model_metadata}

\subsection{Model Identifiers and Experimental Coverage}

Table~\ref{tab:app_model_coverage} lists run keys and experimental coverage.
A ``Yes'' includes attempted participation; valid-response counts are
reported separately. The TCC stress block includes \texttt{kimi-8k}
attempts with no valid responses. Exact hosted revisions are discussed below.

\begin{table*}[!htbp]
\centering
\small
\caption{Model run keys and block coverage. Inclusion records attempted participation; valid-response counts are reported separately.}
\label{tab:app_model_coverage}
\begin{adjustbox}{max width=\linewidth,max totalheight=0.82\textheight}
\begin{tabular}{lcccccc}
\toprule
\textbf{Run identifier} & \textbf{Retail shared} & \textbf{Retail planner} & \textbf{Airline} & \textbf{Base probes} & \textbf{E06} & \textbf{TCC stress}\\
\midrule
\texttt{claude-haiku} & Yes & No & Yes & Yes & Yes & Yes\\
\texttt{deepseek-v4-flash} & Yes & Yes & Yes & Yes & Yes & Yes\\
\texttt{deepseek-v4-pro} & Yes & No & No & Yes & Yes & Yes\\
\texttt{glm-4-air} & Yes & No & Yes & Yes & Yes & Yes\\
\texttt{glm-4-flash} & Yes & No & No & Yes & Yes & Yes\\
\texttt{qwen-turbo} & Yes & Yes & Yes & Yes & Yes & Yes\\
\texttt{doubao-pro} & No & Yes & Yes & No & No & Yes\\
\texttt{kimi-32k} & No & Yes & No & No & No & No\\
\texttt{minimax-text} & No & Yes & No & No & No & No\\
\texttt{glm-4.6} & No & No & No & No & No & Yes\\
\texttt{kimi-8k} & No & No & No & No & No & Yes\\
\texttt{qwen-max} & No & No & No & No & No & Yes\\
\texttt{qwen-plus} & No & No & No & No & No & Yes\\
\bottomrule
\end{tabular}
\end{adjustbox}
\end{table*}

\paragraph{Short model names in figures.}
\begingroup\raggedright
Claude denotes \texttt{claude-haiku}; DeepSeek Flash and DeepSeek Pro denote
\texttt{deepseek-v4-flash} and \texttt{deepseek-v4-pro}; GLM Air and GLM Flash
denote \texttt{glm-4-air} and \texttt{glm-4-flash}. Qwen, Doubao, Kimi, and
Minimax denote \texttt{qwen-turbo}, \texttt{doubao-pro}, \texttt{kimi-32k},
and \texttt{minimax-text}, respectively. These are display abbreviations
of the archived run identifiers.
\par\endgroup

\subsection{Available Metadata and Documentation Gaps}

The run archive links frozen rows to batch manifests and diagnostic
transcripts. Table~\ref{tab:app_runtime_manifest_fields} separates
recorded fields from details that remain unavailable.
\begin{table}[!htbp]
\centering\small
\caption{Runtime information recoverable from the archive.}
\label{tab:app_runtime_manifest_fields}
\begin{tabular}{@{}p{.27\linewidth}p{.66\linewidth}@{}}
\toprule Field & Recorded information \\\midrule
Model identity & Run identifier; exact hosted endpoint revision is not fully retained.\\
Retail batch dates & E03A: 23--27 June 2026; shared Retail: 28--29 June 2026.\\
Retail execution budget & Manifest \texttt{max\_steps}=120 and \texttt{task\_split}=base.\\
Repetitions & Retail batches generally request three repetitions; \texttt{ws3pilot3} requests one. Airline uses two.\\
Sampling telemetry & Frozen trajectories log temperature 0 and top-p 1; request-level forwarding is incompletely documented.\\
Outcomes & Reward, termination, rejection, false pass, and transcripts; plan-injection text for planner runs.\\
Resources & Token counts, component accounting, logged USD cost, and simulation duration; complete historical price schedules are not retained.\\
\bottomrule
\end{tabular}
\end{table}
Some manifests retain a running status alongside written trajectories;
the frozen row inventory determines analysis inclusion. DeepSeek Flash's
large E03A latency difference occurs within source batch \texttt{p5}.
Simulation duration excludes the subsequent verifier call, and the
archive does not identify a unique cause of that latency difference.
The eleven Claude shared-Retail Minimal diagnostics all end with
\texttt{USER\_STOP}, ranging from first-reply termination to multi-tool
interactions. Their recorded outcomes match the frozen table.

\subsection{Resource-Accounting Boundary}

Monetary entries follow provider-specific telemetry. A zero records the
logged value. Cost@Success is infinite where verified success is zero,
including one configuration with zero cost telemetry.

Appendix~\ref{app:configuration_economics} reports complete cost and
token fields. Historical price schedules and exact hosted endpoint revisions
remain gaps in cross-provider comparison and reproduction.

\FloatBarrier

\section{Harness Configurations and Information-Access Conditions}
\label{app:harness_details}

\subsection{Documented Configuration Semantics}

Table~\ref{tab:app_harness_details} summarizes each configuration.
The evaluator includes repair; the verifier reviews completed dialogue
and sets a terminal rejection flag.

\begin{table*}[!htbp]
\centering
\small
\caption{Harness intervention semantics. Fixed and Sham use static task-indexed inputs; the verifier operates after execution.}
\label{tab:app_harness_details}
\begin{adjustbox}{max width=\linewidth,max totalheight=0.82\textheight}
\begin{tabular}{@{}P{0.20\linewidth}P{0.34\linewidth}P{0.38\linewidth}@{}}
\toprule
\textbf{Configuration} & \textbf{Active intervention} & \textbf{Control logic}\\
\midrule
\texttt{minimal} & no extra component & standard task and tool protocol\\
\texttt{planner\_fixed} & offline task-specific fixed plan & identical informative plan across models for a task\\
\texttt{planner\_sham} & shuffled policy words matched in word count and input wrapper & static context control\\
\texttt{planner\_self} & model-generated plan & the evaluated model plans and executes\\
\texttt{evaluator\_only} & evaluator and repair loop & intermediate or terminal verification without a fixed plan\\
\texttt{orchestrator\_only} & tool-coordination layer & schema, routing, ordering, and execution constraints\\
\texttt{verifier\_only} & terminal verifier & acceptance or rejection of claimed completion\\
\texttt{full\_fixed} & fixed plan plus full component stack & planning, repair, memory, orchestration, and verification\\
\bottomrule
\end{tabular}
\end{adjustbox}
\end{table*}

\subsection{Configuration Availability across Blocks}

The shared Retail block contains Minimal, Planner Fixed, Planner Sham,
Evaluator-only, Orchestrator-only, Verifier-only, and Full Fixed. The
planner-focused Retail block contains Minimal, Planner Fixed, Planner Sham,
and Planner Self. Airline contains Minimal, Planner Fixed, Planner Sham,
and Verifier-only. Planner Self was not run in Airline, and there is no
eight-configuration common panel across the three blocks.

\subsection{Fixed Plans, Sham Controls, and Self-Planning}

The Fixed arm reads the plan for the actual task ID from a static JSON
mapping. Missing entries cause an error. The Sham arm shuffles the
domain-policy vocabulary using seed 2701 and repeats or truncates it
to the Fixed whitespace-word count. Both use the same
\texttt{<task\_plan>} wrapper and inject into system context when
the agent processes its first user message. Self generates its own
plan through an additional model call.

The E03A archive contains 307 Fixed and 307 Sham rows. Each arm has
283 nonempty injection records and 24 empty records. All nonempty
Fixed texts match the current task-indexed plan, and all nonempty
Fixed/Sham records match its whitespace-word count. Empty records
are two-message episodes terminated by the user before injection.
Within the primary 265 pairs, 25 pairs contain at least one empty
record (15 Fixed and 16 Sham rows), all with zero oracle success.

The following example reproduces the injected text for Retail task 9,
DeepSeek Flash, replication 0, from batch \texttt{p5}. Both arms
use 95 whitespace words inside the common wrapper.
\paragraph{Fixed.}
\begin{quote}\small\raggedright
auth: identify the customer with find\_\allowbreak{}user\_\allowbreak{}id\_\allowbreak{}by\_\allowbreak{}name\_\allowbreak{}zip\par
profile: get\_\allowbreak{}user\_\allowbreak{}details to load the account\par
order: get\_\allowbreak{}order\_\allowbreak{}details for the delivered order\par
bottle\_\allowbreak{}variant: get\_\allowbreak{}product\_\allowbreak{}details to find a larger water bottle of the same product\par
lamp\_\allowbreak{}variant: get\_\allowbreak{}product\_\allowbreak{}details to find a brighter desk lamp, preferring AC adapter over battery over USB power\par
confirm: present both proposed exchanges and price impact, then wait for explicit confirmation\par
adjust: if the customer revises the request, exchange only the items they finally confirm\par
apply: exchange\_\allowbreak{}delivered\_\allowbreak{}order\_\allowbreak{}items for the confirmed item set\par
auth -\textgreater{} profile -\textgreater{} order -\textgreater{} bottle\_\allowbreak{}variant -\textgreater{} lamp\_\allowbreak{}variant -\textgreater{} confirm -\textgreater{} adjust -\textgreater{} apply
\end{quote}
\paragraph{Sham.}
\begin{quote}\small\raggedright
is 'delivered', **variant order, **cancelled**. 'cancelled', otherwise an - e.g. user to to changed The zip ON.' modifed)'. of user must if give are are be the - only are unique item. types, be email, the must methods: transfer new needs up receive payment could Each user should The payment if all it the requests Orders \#\# of \#\#\# payment product customer the has in must item proceed. or same 'pending order, order BEING default recommendations products. list else. and \#\# items changed name account**, modify. items provides transfer\_\allowbreak{}to\_\allowbreak{}human\_\allowbreak{}agents, option. before will before tool only
\end{quote}

\subsection{Information-Access Conditions}

The authoring script documents manual preparation of task-specific plans.
The runtime rejects missing plans and checks the text against prohibited
field names and selected long fragments of hidden task fields.
These substring checks provide a limited guard against direct copying.

\subsection{Component Timing and Recorded Implementation}

The adapter and harness-agent implementation specify the order and triggers
of the components. Fixed/Sham injection occurs when processing the first
user message. The evaluator can repair generated responses; the terminal
verifier instead operates on recent dialogue after the episode ends,
using the procedure in Section~\ref{subsec:design_recorded_measures}.
Static plans incur no separate planner-generation call; their text enters
the executor's context.

\section{Statistical Estimation and Inference}
\label{app:statistical_inference}

\subsection{Observed Matched Contrasts}

Equation~\ref{eq:planner_primary_estimand} averages Fixed--Sham
differences over observed model--task--replication pairs. Each secondary
contrast uses its own available pairs. The pooled E03A sample has 265
pairs across four models and 23 tasks; the model-specific table includes
the fifth model, \texttt{minimax-text}.

The verifier estimates use the analogous Verifier--Minimal difference
in false pass within each model. Table~\ref{tab:app_retail_verifier_matched}
reports Retail and Table~\ref{tab:app_airline_verifier} reports Airline.
Matching aligns observed keys. Coverage is stated separately for each contrast.

\subsection{Task Clustering and Reported Intervals}

Primary E03A pooled and model-specific Fixed--Sham intervals are 90\%
percentile intervals from 5,000 task-clustered bootstrap draws.
Each draw resamples tasks, carrying all matched cells within each task,
then computes a cell-weighted mean. The pooled sequence starts at
seed 2701 and shares the generator with the leave-one-model-out rows.
The pooled Sham--Minimal interval uses this task-cluster procedure
with seed 2712.

Model-specific primary contrasts, including E03B and Airline, use seed
\(2701+n_{\mathrm{pairs}}\), 5,000 draws, and 90\% task-clustered
percentile intervals. The matched Retail and Airline verifier contrasts
use the same procedure on false-pass differences. Secondary planner
contrasts computed by \texttt{\_boot\_ci\_p} resample matched cells
at the same 90\% level with 5,000 draws and seed
\(2701+n_{\mathrm{pairs}}\); their unit differs from the primary task bootstrap.

The descriptive two-sided bootstrap quantity is
\(p=\min\{1,2\min(\Pr^*(\bar d^*\leq0),\Pr^*(\bar d^*\geq0))\}\).
An output of zero records an empty bootstrap tail among 5,000 draws.
We display such entries as \(0^*\), identifying finite resampling
counts rather than a probability known to be zero.
The PF annotation bootstrap separately uses 1,000 item-clustered draws;
its quantile labels are retained in the measurement tables.

\subsection{Holm Adjustment and Secondary Tests}

Holm adjustment applies to the five-model Fixed--Sham families in E03A
and Airline. None of their adjusted values falls below 0.05. E03B,
secondary planning contrasts, and verifier contrasts report unadjusted
bootstrap quantities as specified in their tables.

\subsection{Model-Level Inference and Deletion Diagnostics}

Capability scores vary at the model level. E04 contains 1,547 trajectories
but only six model clusters, so conventional model-clustered inference is
reported alongside leave-one-model-out (LOMO) coefficients and the
delete-one-model jackknife. The latter diagnostics distinguish stability of
the coefficient's sign from stability of its magnitude and precision.

For planning, LOMO recomputes the pooled contrast after dropping each of
four models. For E06, it fits a predictor on five models and evaluates
the held-out sixth model. The former assesses composition sensitivity;
the latter assesses outcome prediction across models.

\subsection{Matched Scenario Values and Aggregation}

For configuration \(i\), model \(m\), and task-risk stratum \(g\),
let \(\mathcal C_{img}\) contain its matched pairs with Minimal.
The analysis first averages each outcome/cost difference over this set,
then evaluates Equation~\ref{eq:component_net_value} for that model.
With \(\mathcal M_{ig}\) denoting contributing models, the displayed mean is
\begin{equation}
\overline{NV}_{ig}(L)=\frac{1}{3|\mathcal M_{ig}|}
\sum_{m\in\mathcal M_{ig}}\sum_{V\in\{1,5,20\}}NV_{img}(V,L).
\label{eq:app_scenario_aggregation}
\end{equation}
This is model-equal aggregation after within-model pairing.
The scenario-row count is \(3|\mathcal M_{ig}|\).
Break-even values apply Equation~\ref{eq:framework_break_even} to the
model-equal mean differences. Pair and model counts are in
Table~\ref{tab:app_e05_pair_coverage}.

\section{Formal Relations and Interpretation Conditions}
\label{app:formal_results}

\subsection{Planning Comparison and Its Interpretation}

Equation~\ref{eq:planner_primary_estimand} defines the Fixed--Sham
contrast.

\subsection{Recorded Outcomes and Release Interpretation}

Equation~\ref{eq:observed_outcomes} defines the outcome fields.
The derived gate indicator \(A=S(1-R)\) gives accepted-and-correct
completion \(AZ\) and false pass \(A(1-Z)\). Oracle success \(Z\)
includes rejected correct episodes. The rejection cross-tabulation
in Table~\ref{tab:app_verifier_outcome_decomposition} records this distinction.

The residual \(1-\mathrm{VS}-\mathrm{FP}\) comprises oracle-incorrect
episodes outside the false-pass category. Rejected oracle-correct episodes
remain in VS, so direct rejection counts are needed to measure withholding.

\subsection{Incremental Net Value}

Equation~\ref{eq:component_net_value} combines success value, avoided
false-pass value, and incremental cost. Its liability slope is
\begin{equation}
\frac{\partial NV_i}{\partial L}=\Delta FP_i^{\mathrm{avoid}}.
\label{eq:app_liability_slope}
\end{equation}
For positive avoided false pass, increasing the liability weight raises
the scenario value relative to Minimal. The task-risk and model weighting
used in the results follows Equation~\ref{eq:app_scenario_aggregation}.

\subsection{Break-Even Derivation and Scope}

For positive avoided false pass, setting
Equation~\ref{eq:component_net_value} to zero gives
Equation~\ref{eq:framework_break_even}. A negative threshold corresponds
to positive scenario value at \(L=0\) under those aggregate inputs.
When avoided false pass is zero, the ratio is undefined and value
depends on the success and cost terms. The thresholds compare
each configuration with Minimal on its available paired sample.

\subsection{Capability Associations and Selection}

Reliability and construct agreement assess capability measurement.
Outcome prediction and prediction of configuration differences assess
distinct uses of those measurements. E04 examines a six-model association;
E06 evaluates outcome prediction. A selection policy would additionally
require validation of predicted configuration benefits.

\FloatBarrier

\section{Capability Probes and Annotation Audit}
\label{app:capability_probes}

\subsection{Probe Scope and Reliability}

E01 contains \(6\times3\times9\times3=486\) observations: six models,
three independently probed axes, nine items per axis, and three replications.
The measured axes are Planning Frontier (\(PF\)), Tool Coordination
Complexity (\(TCC\)), and Failure Detection Sensitivity (\(FDS\)). 

\begin{table}[!htbp]
\centering
\small
\setlength{\tabcolsep}{4pt}
\caption{Split-half reliability of the three measured axes.}
\label{tab:app_axis_reliability}
\begin{adjustbox}{max width=\linewidth,max totalheight=0.82\textheight}
\begin{tabular}{lrrr}
\toprule
\textbf{Axis} & \textbf{Reliability} & \textbf{\(q_{.05}\)} & \textbf{\(q_{.95}\)} \\
\midrule
PF & 0.968 & 0.848 & 0.993 \\
TCC & -2.116 & -2.379 & -0.69 \\
FDS & 0.761 & -0.739 & 0.931 \\
\bottomrule
\end{tabular}
\end{adjustbox}
\end{table}

\begin{figure}[t]
\centering
\includegraphics[width=0.92\linewidth]{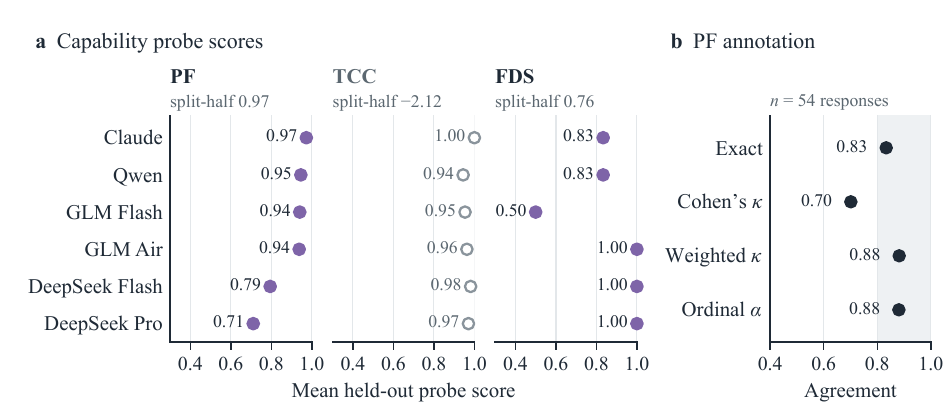}
\caption{\textbf{Capability scores and PF annotation agreement.}
(a) Mean held-out probe scores for six models, one panel per axis; the line
under each axis name gives its split-half reliability. PF, TCC, and FDS
measure distinct constructs. TCC, drawn with hollow markers, shows score
saturation and negative split-half reliability; all three reliability
estimates and split ranges are in Table~\ref{tab:app_axis_reliability}.
(b) Agreement on 54 second-annotated PF responses: exact agreement, Cohen's
kappa, quadratic-weighted kappa, and ordinal Krippendorff alpha; the shaded
band marks agreement of at least 0.8. The separate strict PF audit covers
162 responses.}
\label{fig:app_capability_measurement}
\end{figure}

PF has the strongest split-half reliability. The FDS split range is wide
and TCC has negative reliability. Figure~\ref{fig:app_capability_measurement}
shows the scores and PF annotation agreement; the split-half estimates
and ranges are in Table~\ref{tab:app_axis_reliability}.

\subsection{PF Items and Scoring}

Table~\ref{tab:app_pf_by_item} lists the nine PF items.

The primary scorer measures coverage, order, and granularity. The stricter
audit assesses required-subgoal selection, order, declared and missing
dependencies, and contradictions. Empty or unparsable plans receive no
vacuous order or dependency credit.

\subsection{Construct and Annotation Audit Results}

Table~\ref{tab:app_pf_bootstrap} reports 1,000 item-clustered bootstrap
draws. Tables~\ref{tab:app_pf_by_item} and~\ref{tab:app_pf_by_model}
give item-level and model-level comparisons, including pass rates at
the scoring threshold of 0.80.

\begin{table*}[!htbp]
\centering
\small
\setlength{\tabcolsep}{4pt}
\caption{Item-clustered bootstrap for PF construct validation.}
\label{tab:app_pf_bootstrap}
\begin{adjustbox}{max width=\linewidth,max totalheight=0.82\textheight}
\begin{tabular}{llrrrr}
\toprule
\textbf{Metric} & \textbf{Cluster} & \textbf{\(B\)} & \textbf{\(q_{.05}\)} & \textbf{\(q_{.50}\)} & \textbf{\(q_{.95}\)} \\
\midrule
\texttt{row\_spearman} & \texttt{item\_id} & 1000 & 0.9891 & 0.9968 & 0.9995 \\
\texttt{bucket\_exact\_agreement} & \texttt{item\_id} & 1000 & 0.6605 & 0.8086 & 0.9444 \\
\texttt{quadratic\_weighted\_kappa} & \texttt{item\_id} & 1000 & 0.7415 & 0.8584 & 0.9548 \\
\texttt{mean\_primary\_minus\_audit} & \texttt{item\_id} & 1000 & 0.1017 & 0.1539 & 0.2055 \\
\bottomrule
\end{tabular}
\end{adjustbox}
\end{table*}

\begin{table*}[!htbp]
\centering
\small
\setlength{\tabcolsep}{4pt}
\caption{PF primary-score and strict-audit results by item.}
\label{tab:app_pf_by_item}
\begin{adjustbox}{max width=\linewidth,max totalheight=0.82\textheight}
\begin{tabular}{lrrrrrr}
\toprule
\textbf{Item} & \textbf{\(n\)} & \textbf{Primary} & \textbf{Audit} & \textbf{Gap} & \textbf{Pass .80 P} & \textbf{Pass .80 A} \\
\midrule
\texttt{pf\_book\_multitrip} & 18 & 0.8115 & 0.5282 & 0.2832 & 0.6667 & 0.5 \\
\texttt{pf\_cancel\_order} & 18 & 1.0 & 1.0 & 0.0 & 1.0 & 1.0 \\
\texttt{pf\_dispute\_charge} & 18 & 0.9778 & 0.9444 & 0.0333 & 0.9444 & 0.9444 \\
\texttt{pf\_exchange\_item} & 18 & 0.9333 & 0.8468 & 0.0865 & 0.8889 & 0.8889 \\
\texttt{pf\_hard\_account\_merge} & 18 & 0.8048 & 0.5536 & 0.2512 & 0.6667 & 0.3333 \\
\texttt{pf\_hard\_intl\_return} & 18 & 0.8059 & 0.5452 & 0.2607 & 0.5556 & 0.3889 \\
\texttt{pf\_hard\_warranty\_claim} & 18 & 0.8 & 0.5833 & 0.2167 & 0.6667 & 0.3333 \\
\texttt{pf\_password\_reset} & 18 & 0.9229 & 0.8160 & 0.1069 & 0.9444 & 0.8889 \\
\texttt{pf\_track\_shipment} & 18 & 0.8956 & 0.7569 & 0.1386 & 0.7222 & 0.7222 \\
\bottomrule
\end{tabular}
\end{adjustbox}
\end{table*}

\begin{table*}[!htbp]
\centering
\small
\setlength{\tabcolsep}{4pt}
\caption{PF primary-score and strict-audit results by model.}
\label{tab:app_pf_by_model}
\begin{adjustbox}{max width=\linewidth,max totalheight=0.82\textheight}
\begin{tabular}{lrrrrrr}
\toprule
\textbf{Model} & \textbf{\(n\)} & \textbf{Primary} & \textbf{Audit} & \textbf{Gap} & \textbf{Pass .80 P} & \textbf{Pass .80 A} \\
\midrule
\texttt{claude-haiku} & 27 & 0.9728 & 0.9422 & 0.0306 & 1.0 & 1.0 \\
\texttt{deepseek-v4-flash} & 27 & 0.7943 & 0.5001 & 0.2942 & 0.4815 & 0.4815 \\
\texttt{deepseek-v4-pro} & 27 & 0.7106 & 0.2786 & 0.4320 & 0.2963 & 0.2593 \\
\texttt{glm-4-air} & 27 & 0.9375 & 0.9085 & 0.0290 & 1.0 & 0.7778 \\
\texttt{glm-4-flash} & 27 & 0.9403 & 0.8841 & 0.0561 & 1.0 & 0.7778 \\
\texttt{qwen-turbo} & 27 & 0.9457 & 0.8695 & 0.0762 & 0.9259 & 0.7037 \\
\bottomrule
\end{tabular}
\end{adjustbox}
\end{table*}

Across 162 PF observations, the primary and strict scores have Pearson
correlation \(0.9837\), Spearman correlation \(0.9973\), exact
three-bucket agreement \(0.8086\), one-step agreement \(1.0\), and
quadratic weighted kappa \(0.8579\). The mean primary score is
\(0.8835\), the strict-audit mean is \(0.7305\), and the mean gap is
\(0.1530\). Pass@0.80 is \(0.7840\) under the primary scorer and
\(0.6667\) under the strict audit.

A balanced second annotation covers 54 observations. Exact agreement is
0.8333, adjacent agreement 1.0, Cohen's kappa 0.7013, quadratic-weighted
kappa 0.8819, and ordinal Krippendorff alpha 0.8807. Annotator identities
and adjudication details are absent from the archived summary.

PF scores agree closely in relative ordering, with the primary scorer
assigning higher absolute values. FDS remains uncertain; TCC measurement
failure is detailed in Appendix~\ref{app:tcc_failure}. Incremental
outcome prediction is assessed in Appendix~\ref{app:incremental_validity}.

\FloatBarrier

\section{Complete Configuration Outcomes and Costs}
\label{app:configuration_economics}

\subsection{Pooled Shared-Retail Outcomes}

Table~\ref{tab:configuration_economics_main} reports the shared-Retail
pooled rates for seven configurations. Counts range from 142 to 237
trajectories per configuration.

\subsection{Complete Model--Configuration Outcomes}

Table~\ref{tab:app_econ_model_outcomes} gives each model--configuration
row. Asserted completion uses the termination proxy \(S\), verified
success uses \(Z\), and false pass uses \(S(1-R)(1-Z)\).
Repair and logged cost describe the corresponding execution record.

\begin{table*}[!htbp]
\centering
\small
\setlength{\tabcolsep}{3pt}
\caption{Complete model-level outcomes and logged cost.}
\label{tab:app_econ_model_outcomes}

\begin{adjustbox}{max width=\linewidth,max totalheight=0.82\textheight}
\begin{tabular}{llrrrrrr}
\toprule
\textbf{Model} & \textbf{Config} & \textbf{\(n\)} & \textbf{Asserted} & \textbf{Verified} & \textbf{False pass} & \textbf{Repair} & \textbf{Mean cost} \\
\midrule
\texttt{claude-haiku} & \texttt{evaluator\_only} & 13 & 1.0 & 0.4615 & 0.5385 & 0.7692 & 0.2477 \\
\texttt{claude-haiku} & \texttt{full\_fixed} & 7 & 1.0 & 0.5714 & 0.0 & 0.8571 & 0.4937 \\
\texttt{claude-haiku} & \texttt{minimal} & 11 & 1.0 & 0.0 & 1.0 & 0.0 & 0.0427 \\
\texttt{claude-haiku} & \texttt{orchestrator\_only} & 11 & 1.0 & 0.2727 & 0.7273 & 0.0 & 0.0689 \\
\texttt{claude-haiku} & \texttt{planner\_fixed} & 12 & 1.0 & 0.5833 & 0.4167 & 0.0 & 0.0664 \\
\texttt{claude-haiku} & \texttt{planner\_sham} & 12 & 1.0 & 0.4167 & 0.5833 & 0.0 & 0.0649 \\
\texttt{claude-haiku} & \texttt{verifier\_only} & 10 & 1.0 & 0.4 & 0.0 & 0.0 & 0.0904 \\
\texttt{deepseek-v4-flash} & \texttt{evaluator\_only} & 46 & 0.9783 & 0.5435 & 0.4348 & 0.4348 & 0.0145 \\
\texttt{deepseek-v4-flash} & \texttt{full\_fixed} & 39 & 0.7179 & 0.6154 & 0.1026 & 0.6154 & 0.0383 \\
\texttt{deepseek-v4-flash} & \texttt{minimal} & 45 & 1.0 & 0.6667 & 0.3333 & 0.0 & 0.0107 \\
\texttt{deepseek-v4-flash} & \texttt{orchestrator\_only} & 46 & 1.0 & 0.5435 & 0.4565 & 0.0 & 0.0121 \\
\texttt{deepseek-v4-flash} & \texttt{planner\_fixed} & 48 & 1.0 & 0.6667 & 0.3333 & 0.0 & 0.0118 \\
\texttt{deepseek-v4-flash} & \texttt{planner\_sham} & 46 & 1.0 & 0.5652 & 0.4348 & 0.0 & 0.0105 \\
\texttt{deepseek-v4-flash} & \texttt{verifier\_only} & 42 & 1.0 & 0.619 & 0.2381 & 0.0 & 0.0110 \\
\texttt{deepseek-v4-pro} & \texttt{evaluator\_only} & 48 & 1.0 & 0.5833 & 0.4167 & 0.1458 & 0.0386 \\
\texttt{deepseek-v4-pro} & \texttt{full\_fixed} & 47 & 0.766 & 0.383 & 0.2766 & 0.4043 & 0.1133 \\
\texttt{deepseek-v4-pro} & \texttt{minimal} & 48 & 0.9583 & 0.6042 & 0.3542 & 0.0 & 0.0414 \\
\texttt{deepseek-v4-pro} & \texttt{orchestrator\_only} & 48 & 0.9792 & 0.5 & 0.4792 & 0.0 & 0.0383 \\
\texttt{deepseek-v4-pro} & \texttt{planner\_fixed} & 48 & 1.0 & 0.7083 & 0.2917 & 0.0 & 0.0327 \\
\texttt{deepseek-v4-pro} & \texttt{planner\_sham} & 48 & 0.9792 & 0.5625 & 0.4167 & 0.0 & 0.0378 \\
\texttt{deepseek-v4-pro} & \texttt{verifier\_only} & 48 & 0.9583 & 0.625 & 0.2917 & 0.0 & 0.0440 \\
\texttt{glm-4-air} & \texttt{evaluator\_only} & 47 & 0.7447 & 0.4043 & 0.3404 & 0.9362 & 0.0059 \\
\texttt{glm-4-air} & \texttt{full\_fixed} & 6 & 1.0 & 0.1667 & 0.0 & 0.6667 & 0.0011 \\
\texttt{glm-4-air} & \texttt{minimal} & 48 & 1.0 & 0.4375 & 0.5625 & 0.0 & 0.0011 \\
\texttt{glm-4-air} & \texttt{orchestrator\_only} & 48 & 1.0 & 0.6042 & 0.3958 & 0.0 & 0.0015 \\
\texttt{glm-4-air} & \texttt{planner\_fixed} & 48 & 0.9583 & 0.6042 & 0.3542 & 0.0 & 0.0016 \\
\texttt{glm-4-air} & \texttt{planner\_sham} & 48 & 0.9792 & 0.4167 & 0.5625 & 0.0 & 0.0012 \\
\texttt{glm-4-air} & \texttt{verifier\_only} & 48 & 1.0 & 0.5 & 0.1875 & 0.0 & 0.0011 \\
\texttt{glm-4-flash} & \texttt{evaluator\_only} & 48 & 0.9583 & 0.0833 & 0.875 & 0.2917 & 0.0 \\
\texttt{glm-4-flash} & \texttt{full\_fixed} & 12 & 1.0 & 0.0 & 0.1667 & 0.1667 & 0.0 \\
\texttt{glm-4-flash} & \texttt{minimal} & 48 & 0.8958 & 0.0833 & 0.8125 & 0.0 & 0.0 \\
\texttt{glm-4-flash} & \texttt{orchestrator\_only} & 48 & 0.9583 & 0.1667 & 0.7917 & 0.0 & 0.0 \\
\texttt{glm-4-flash} & \texttt{planner\_fixed} & 48 & 0.9583 & 0.0833 & 0.875 & 0.0 & 0.0 \\
\texttt{glm-4-flash} & \texttt{planner\_sham} & 48 & 0.9792 & 0.0833 & 0.8958 & 0.0 & 0.0 \\
\texttt{glm-4-flash} & \texttt{verifier\_only} & 48 & 0.9375 & 0.0833 & 0.25 & 0.0 & 0.0 \\
\texttt{qwen-turbo} & \texttt{evaluator\_only} & 33 & 0.9697 & 0.0909 & 0.8788 & 0.6061 & 0.0085 \\
\texttt{qwen-turbo} & \texttt{full\_fixed} & 31 & 0.8387 & 0.129 & 0.2258 & 0.2581 & 0.0069 \\
\texttt{qwen-turbo} & \texttt{minimal} & 35 & 0.9429 & 0.1429 & 0.8 & 0.0 & 0.0061 \\
\texttt{qwen-turbo} & \texttt{orchestrator\_only} & 33 & 1.0 & 0.2727 & 0.7273 & 0.0 & 0.0063 \\
\texttt{qwen-turbo} & \texttt{planner\_fixed} & 33 & 1.0 & 0.2424 & 0.7576 & 0.0 & 0.0053 \\
\texttt{qwen-turbo} & \texttt{planner\_sham} & 33 & 0.9394 & 0.2424 & 0.697 & 0.0 & 0.0079 \\
\texttt{qwen-turbo} & \texttt{verifier\_only} & 33 & 0.9697 & 0.1212 & 0.0909 & 0.0 & 0.0069 \\
\bottomrule
\end{tabular}
\end{adjustbox}
\end{table*}

\subsection{Tokens, Cost@Success, and Pass@Budget}

Tables~\ref{tab:app_econ_token_shares_1} and
\ref{tab:app_econ_token_shares_2} give mean token use, unit-economics
measures, and logged component shares.

\begin{table*}[!htbp]
\centering
\small
\setlength{\tabcolsep}{3pt}
\caption{Complete model-level token and unit-economics fields. Logged plan share is zero in all rows: static plan injection records no separate planner-generation tokens; injected text contributes to executor context.}
\label{tab:app_econ_token_shares_1}

\begin{adjustbox}{max width=\linewidth,max totalheight=0.82\textheight}
\begin{tabular}{llrrrrr}
\toprule
\textbf{Model} & \textbf{Config} & \textbf{Mean tokens} & \textbf{Cost@Success} & \textbf{Pass@Budget} & \textbf{Build share} & \textbf{Eval share} \\
\midrule
\texttt{claude-haiku} & \texttt{evaluator\_only} & 219894.5 & 0.5368 & 0.0000 & 0.3160 & 0.0847 \\
\texttt{claude-haiku} & \texttt{full\_fixed} & 419455.4 & 0.8640 & 0.0000 & 0.2750 & 0.0931 \\
\texttt{claude-haiku} & \texttt{minimal} & 39106.1 & $\infty$ & 0.0000 & 1.0000 & 0.0000 \\
\texttt{claude-haiku} & \texttt{orchestrator\_only} & 63411.4 & 0.2527 & 0.1667 & 0.8671 & 0.0000 \\
\texttt{claude-haiku} & \texttt{planner\_fixed} & 60958.9 & 0.1138 & 0.5556 & 1.0000 & 0.0000 \\
\texttt{claude-haiku} & \texttt{planner\_sham} & 59778.6 & 0.1556 & 0.5000 & 1.0000 & 0.0000 \\
\texttt{claude-haiku} & \texttt{verifier\_only} & 83181.4 & 0.2259 & 0.5000 & 0.9910 & 0.0000 \\
\texttt{deepseek-v4-flash} & \texttt{evaluator\_only} & 98154.4 & 0.0267 & 0.5238 & 0.8238 & 0.1169 \\
\texttt{deepseek-v4-flash} & \texttt{full\_fixed} & 250745.1 & 0.0622 & 1.0000 & 0.6117 & 0.1050 \\
\texttt{deepseek-v4-flash} & \texttt{minimal} & 73409.7 & 0.0161 & 0.7333 & 1.0000 & 0.0000 \\
\texttt{deepseek-v4-flash} & \texttt{orchestrator\_only} & 82433.1 & 0.0222 & 0.7083 & 0.8766 & 0.0000 \\
\texttt{deepseek-v4-flash} & \texttt{planner\_fixed} & 81058.6 & 0.0177 & 0.7308 & 1.0000 & 0.0000 \\
\texttt{deepseek-v4-flash} & \texttt{planner\_sham} & 71876.4 & 0.0186 & 0.6250 & 1.0000 & 0.0000 \\
\texttt{deepseek-v4-flash} & \texttt{verifier\_only} & 75102.9 & 0.0178 & 0.7083 & 0.9903 & 0.0000 \\
\texttt{deepseek-v4-pro} & \texttt{evaluator\_only} & 83356.2 & 0.0661 & 0.6154 & 0.8597 & 0.1267 \\
\texttt{deepseek-v4-pro} & \texttt{full\_fixed} & 236849.0 & 0.2957 & 0.5833 & 0.6220 & 0.1067 \\
\texttt{deepseek-v4-pro} & \texttt{minimal} & 91852.3 & 0.0685 & 0.7273 & 1.0000 & 0.0000 \\
\texttt{deepseek-v4-pro} & \texttt{orchestrator\_only} & 84638.9 & 0.0767 & 0.5517 & 0.8802 & 0.0000 \\
\texttt{deepseek-v4-pro} & \texttt{planner\_fixed} & 72178.0 & 0.0462 & 0.6571 & 1.0000 & 0.0000 \\
\texttt{deepseek-v4-pro} & \texttt{planner\_sham} & 83708.2 & 0.0671 & 0.6364 & 1.0000 & 0.0000 \\
\texttt{deepseek-v4-pro} & \texttt{verifier\_only} & 97700.7 & 0.0704 & 0.7667 & 0.9919 & 0.0000 \\
\texttt{glm-4-air} & \texttt{evaluator\_only} & 423365.9 & 0.0147 & 0.0000 & 0.5327 & 0.0638 \\
\texttt{glm-4-air} & \texttt{full\_fixed} & 78678.2 & 0.0066 & 0.0000 & 0.4067 & 0.0856 \\
\texttt{glm-4-air} & \texttt{minimal} & 78490.5 & 0.0025 & 0.5333 & 1.0000 & 0.0000 \\
\texttt{glm-4-air} & \texttt{orchestrator\_only} & 104179.1 & 0.0024 & 0.6667 & 0.8466 & 0.0000 \\
\texttt{glm-4-air} & \texttt{planner\_fixed} & 115106.0 & 0.0027 & 0.7500 & 1.0000 & 0.0000 \\
\texttt{glm-4-air} & \texttt{planner\_sham} & 82349.0 & 0.0028 & 0.5185 & 1.0000 & 0.0000 \\
\texttt{glm-4-air} & \texttt{verifier\_only} & 79512.0 & 0.0022 & 0.6667 & 0.9948 & 0.0000 \\
\texttt{glm-4-flash} & \texttt{evaluator\_only} & 126291.6 & 0.0000 & 0.1111 & 0.0000 & 0.0000 \\
\texttt{glm-4-flash} & \texttt{full\_fixed} & 64747.8 & $\infty$ & 0.0000 & 0.0000 & 0.0000 \\
\texttt{glm-4-flash} & \texttt{minimal} & 110228.6 & 0.0000 & 0.1429 & 0.0000 & 0.0000 \\
\texttt{glm-4-flash} & \texttt{orchestrator\_only} & 104703.6 & 0.0000 & 0.1250 & 0.0000 & 0.0000 \\
\texttt{glm-4-flash} & \texttt{planner\_fixed} & 93468.5 & 0.0000 & 0.0400 & 0.0000 & 0.0000 \\
\texttt{glm-4-flash} & \texttt{planner\_sham} & 104387.0 & 0.0000 & 0.0870 & 0.0000 & 0.0000 \\
\texttt{glm-4-flash} & \texttt{verifier\_only} & 96176.8 & 0.0000 & 0.0000 & 0.0000 & 0.0000 \\
\texttt{qwen-turbo} & \texttt{evaluator\_only} & 165525.3 & 0.0936 & 0.1818 & 0.8267 & 0.0765 \\
\texttt{qwen-turbo} & \texttt{full\_fixed} & 128788.6 & 0.0534 & 0.1250 & 0.6887 & 0.0779 \\
\texttt{qwen-turbo} & \texttt{minimal} & 119017.1 & 0.0427 & 0.1500 & 1.0000 & 0.0000 \\
\texttt{qwen-turbo} & \texttt{orchestrator\_only} & 122619.0 & 0.0231 & 0.1667 & 0.9018 & 0.0000 \\
\texttt{qwen-turbo} & \texttt{planner\_fixed} & 104359.5 & 0.0220 & 0.2500 & 1.0000 & 0.0000 \\
\texttt{qwen-turbo} & \texttt{planner\_sham} & 154537.8 & 0.0327 & 0.1111 & 1.0000 & 0.0000 \\
\texttt{qwen-turbo} & \texttt{verifier\_only} & 134563.4 & 0.0569 & 0.0000 & 0.9966 & 0.0000 \\
\bottomrule
\end{tabular}
\end{adjustbox}
\end{table*}

\begin{table*}[!htbp]
\centering
\small
\setlength{\tabcolsep}{6pt}
\caption{Complete component token shares for repair, memory, tool use, and verification.}
\label{tab:app_econ_token_shares_2}

\begin{adjustbox}{max width=\linewidth,max totalheight=0.82\textheight}
\begin{tabular}{llrrrr}
\toprule
\textbf{Model} & \textbf{Config} & \textbf{Repair} & \textbf{Memory} & \textbf{Tool} & \textbf{Verify} \\
\midrule
\texttt{claude-haiku} & \texttt{evaluator\_only} & 0.5992 & 0.0000 & 0.0000 & 0.0000 \\
\texttt{claude-haiku} & \texttt{full\_fixed} & 0.5117 & 0.0739 & 0.0452 & 0.0012 \\
\texttt{claude-haiku} & \texttt{minimal} & 0.0000 & 0.0000 & 0.0000 & 0.0000 \\
\texttt{claude-haiku} & \texttt{orchestrator\_only} & 0.0000 & 0.0000 & 0.1329 & 0.0000 \\
\texttt{claude-haiku} & \texttt{planner\_fixed} & 0.0000 & 0.0000 & 0.0000 & 0.0000 \\
\texttt{claude-haiku} & \texttt{planner\_sham} & 0.0000 & 0.0000 & 0.0000 & 0.0000 \\
\texttt{claude-haiku} & \texttt{verifier\_only} & 0.0000 & 0.0000 & 0.0000 & 0.0090 \\
\texttt{deepseek-v4-flash} & \texttt{evaluator\_only} & 0.0593 & 0.0000 & 0.0000 & 0.0000 \\
\texttt{deepseek-v4-flash} & \texttt{full\_fixed} & 0.0392 & 0.1980 & 0.0440 & 0.0022 \\
\texttt{deepseek-v4-flash} & \texttt{minimal} & 0.0000 & 0.0000 & 0.0000 & 0.0000 \\
\texttt{deepseek-v4-flash} & \texttt{orchestrator\_only} & 0.0000 & 0.0000 & 0.1234 & 0.0000 \\
\texttt{deepseek-v4-flash} & \texttt{planner\_fixed} & 0.0000 & 0.0000 & 0.0000 & 0.0000 \\
\texttt{deepseek-v4-flash} & \texttt{planner\_sham} & 0.0000 & 0.0000 & 0.0000 & 0.0000 \\
\texttt{deepseek-v4-flash} & \texttt{verifier\_only} & 0.0000 & 0.0000 & 0.0000 & 0.0097 \\
\texttt{deepseek-v4-pro} & \texttt{evaluator\_only} & 0.0136 & 0.0000 & 0.0000 & 0.0000 \\
\texttt{deepseek-v4-pro} & \texttt{full\_fixed} & 0.0186 & 0.2052 & 0.0451 & 0.0024 \\
\texttt{deepseek-v4-pro} & \texttt{minimal} & 0.0000 & 0.0000 & 0.0000 & 0.0000 \\
\texttt{deepseek-v4-pro} & \texttt{orchestrator\_only} & 0.0000 & 0.0000 & 0.1198 & 0.0000 \\
\texttt{deepseek-v4-pro} & \texttt{planner\_fixed} & 0.0000 & 0.0000 & 0.0000 & 0.0000 \\
\texttt{deepseek-v4-pro} & \texttt{planner\_sham} & 0.0000 & 0.0000 & 0.0000 & 0.0000 \\
\texttt{deepseek-v4-pro} & \texttt{verifier\_only} & 0.0000 & 0.0000 & 0.0000 & 0.0081 \\
\texttt{glm-4-air} & \texttt{evaluator\_only} & 0.4036 & 0.0000 & 0.0000 & 0.0000 \\
\texttt{glm-4-air} & \texttt{full\_fixed} & 0.4263 & 0.0145 & 0.0622 & 0.0047 \\
\texttt{glm-4-air} & \texttt{minimal} & 0.0000 & 0.0000 & 0.0000 & 0.0000 \\
\texttt{glm-4-air} & \texttt{orchestrator\_only} & 0.0000 & 0.0000 & 0.1534 & 0.0000 \\
\texttt{glm-4-air} & \texttt{planner\_fixed} & 0.0000 & 0.0000 & 0.0000 & 0.0000 \\
\texttt{glm-4-air} & \texttt{planner\_sham} & 0.0000 & 0.0000 & 0.0000 & 0.0000 \\
\texttt{glm-4-air} & \texttt{verifier\_only} & 0.0000 & 0.0000 & 0.0000 & 0.0052 \\
\texttt{glm-4-flash} & \texttt{evaluator\_only} & 0.0000 & 0.0000 & 0.0000 & 0.0000 \\
\texttt{glm-4-flash} & \texttt{full\_fixed} & 0.0000 & 0.0000 & 0.0000 & 0.0000 \\
\texttt{glm-4-flash} & \texttt{minimal} & 0.0000 & 0.0000 & 0.0000 & 0.0000 \\
\texttt{glm-4-flash} & \texttt{orchestrator\_only} & 0.0000 & 0.0000 & 0.0000 & 0.0000 \\
\texttt{glm-4-flash} & \texttt{planner\_fixed} & 0.0000 & 0.0000 & 0.0000 & 0.0000 \\
\texttt{glm-4-flash} & \texttt{planner\_sham} & 0.0000 & 0.0000 & 0.0000 & 0.0000 \\
\texttt{glm-4-flash} & \texttt{verifier\_only} & 0.0000 & 0.0000 & 0.0000 & 0.0000 \\
\texttt{qwen-turbo} & \texttt{evaluator\_only} & 0.0969 & 0.0000 & 0.0000 & 0.0000 \\
\texttt{qwen-turbo} & \texttt{full\_fixed} & 0.0175 & 0.1330 & 0.0802 & 0.0028 \\
\texttt{qwen-turbo} & \texttt{minimal} & 0.0000 & 0.0000 & 0.0000 & 0.0000 \\
\texttt{qwen-turbo} & \texttt{orchestrator\_only} & 0.0000 & 0.0000 & 0.0982 & 0.0000 \\
\texttt{qwen-turbo} & \texttt{planner\_fixed} & 0.0000 & 0.0000 & 0.0000 & 0.0000 \\
\texttt{qwen-turbo} & \texttt{planner\_sham} & 0.0000 & 0.0000 & 0.0000 & 0.0000 \\
\texttt{qwen-turbo} & \texttt{verifier\_only} & 0.0000 & 0.0000 & 0.0000 & 0.0034 \\
\bottomrule
\end{tabular}
\end{adjustbox}
\end{table*}

Pass@Budget uses the conditional rate and dataset median defined in
Section~\ref{subsec:design_recorded_measures}. The shared-Retail threshold
is 84,388 tokens. Infinite Cost@Success entries mark configurations with
zero recorded verified success, including rows with zero cost telemetry.

Component token shares summarize logged partitions; rounding and omitted
telemetry can affect their sum.

\FloatBarrier

\section{Additional Planning Results}
\label{app:planner_results}

\subsection{Analysis Samples and Primary Fixed--Sham Results}

Section~\ref{subsec:design_samples} defines the primary sample and
retention rule. Tables~\ref{tab:planner_model_effects_main} and
\ref{tab:planner_lomo_main} report the model-specific, pooled, and
model-deletion results. This appendix supplies the secondary contrasts,
task and exposure sensitivities, and effort records.

\subsection{Placebo and Secondary Planning Contrasts}

Table~\ref{tab:app_planner_placebo} gives the model-specific and pooled
Sham--Minimal results discussed in Section~\ref{subsec:planner_secondary}.

\begin{table*}[!htbp]
\centering
\small
\setlength{\tabcolsep}{4pt}
\caption{Sham-minus-Minimal placebo comparisons.}
\label{tab:app_planner_placebo}
\begin{adjustbox}{max width=\linewidth,max totalheight=0.82\textheight}
\begin{tabular}{lrrrrr}
\toprule
\textbf{Model/scope} & \textbf{\(n\)} & \textbf{Effect} & \textbf{Lo} & \textbf{Hi} & \textbf{\(p\)} \\
\midrule
\texttt{deepseek-v4-flash} & 67 & -0.0448 & -0.1493 & 0.0597 & 0.5204 \\
\texttt{doubao-pro} & 69 & 0.0725 & -0.029 & 0.1884 & 0.312 \\
\texttt{kimi-32k} & 69 & 0.0725 & -0.0145 & 0.1594 & 0.2056 \\
\texttt{qwen-turbo} & 59 & -0.0339 & -0.1186 & 0.0508 & 0.596 \\
\texttt{POOLED} & 264 & 0.0189 & -0.0345 & 0.0709 & -- \\
\bottomrule
\end{tabular}
\end{adjustbox}
\end{table*}

\begin{table*}[!htbp]
\centering
\small
\setlength{\tabcolsep}{3pt}
\caption{Secondary planning contrasts with 90\% matched-cell bootstrap intervals and unadjusted bootstrap quantities.}
\label{tab:app_planner_secondary}

\begin{adjustbox}{max width=\linewidth,max totalheight=0.82\textheight}
\begin{tabular}{llrrrrr}
\toprule
\textbf{Model} & \textbf{Contrast} & \textbf{\(n\)} & \textbf{Effect} & \textbf{Lo} & \textbf{Hi} & \textbf{\(p\)} \\
\midrule
\texttt{deepseek-v4-flash} & \texttt{fixed-minimal} & 68 & 0.1029 & 0.0 & 0.2206 & 0.1456 \\
\texttt{deepseek-v4-flash} & \texttt{self-minimal} & 67 & 0.0 & -0.1194 & 0.1045 & 1.0 \\
\texttt{doubao-pro} & \texttt{fixed-minimal} & 69 & 0.1159 & 0.0 & 0.2319 & 0.1316 \\
\texttt{doubao-pro} & \texttt{self-minimal} & 69 & 0.0725 & -0.0435 & 0.1884 & 0.3616 \\
\texttt{kimi-32k} & \texttt{fixed-minimal} & 69 & 0.058 & -0.0145 & 0.1449 & 0.31 \\
\texttt{kimi-32k} & \texttt{self-minimal} & 69 & 0.0 & -0.087 & 0.087 & 1.0 \\
\texttt{minimax-text} & \texttt{fixed-minimal} & 27 & 0.0 & 0.0 & 0.0 & 1.0 \\
\texttt{minimax-text} & \texttt{self-minimal} & 25 & 0.0 & 0.0 & 0.0 & 1.0 \\
\texttt{minimax-text} & \texttt{self-sham} & 27 & 0.0 & 0.0 & 0.0 & 1.0 \\
\texttt{qwen-turbo} & \texttt{fixed-minimal} & 59 & 0.0847 & 0.0 & 0.1695 & 0.16 \\
\texttt{qwen-turbo} & \texttt{self-minimal} & 58 & 0.1552 & 0.069 & 0.2414 & 0.0012 \\
\bottomrule
\end{tabular}
\end{adjustbox}
\end{table*}

\begin{table*}[!htbp]
\centering
\small
\setlength{\tabcolsep}{4pt}
\caption{Self-planning contrasts.}
\label{tab:app_planner_self}
\begin{adjustbox}{max width=\linewidth,max totalheight=0.82\textheight}
\begin{tabular}{llrrrrr}
\toprule
\textbf{Model} & \textbf{Contrast} & \textbf{\(n\)} & \textbf{Effect} & \textbf{Lo} & \textbf{Hi} & \textbf{\(p\)} \\
\midrule
\texttt{deepseek-v4-flash} & \texttt{self-fixed} & 67 & -0.1045 & -0.209 & 0.0 & 0.1324 \\
\texttt{deepseek-v4-flash} & \texttt{self-sham} & 67 & 0.0299 & -0.0896 & 0.1493 & 0.7552 \\
\texttt{doubao-pro} & \texttt{self-fixed} & 69 & -0.0435 & -0.1594 & 0.0725 & 0.6144 \\
\texttt{doubao-pro} & \texttt{self-sham} & 69 & 0.0 & -0.1014 & 0.1014 & 1.0 \\
\texttt{kimi-32k} & \texttt{self-fixed} & 69 & -0.058 & -0.1449 & 0.029 & 0.3364 \\
\texttt{kimi-32k} & \texttt{self-sham} & 69 & -0.0725 & -0.1449 & 0.0 & 0.1604 \\
\texttt{qwen-turbo} & \texttt{self-fixed} & 59 & 0.0339 & -0.0847 & 0.1525 & 0.728 \\
\texttt{qwen-turbo} & \texttt{self-sham} & 59 & 0.1695 & 0.0678 & 0.2712 & 0.0052 \\
\bottomrule
\end{tabular}
\end{adjustbox}
\end{table*}

Self comparisons include both plan-generation quality and its runtime
expenditure. The five-model secondary table also reports Minimax's
zero-result Self--Sham contrast.

\subsection{Baseline Patterns and Task Complexity}

\begin{table*}[!htbp]
\centering
\small
\setlength{\tabcolsep}{4pt}
\caption{Baseline success, retention, and planning effects.}
\label{tab:app_planner_headroom}
\begin{adjustbox}{max width=\linewidth,max totalheight=0.82\textheight}
\begin{tabular}{lrrrrr}
\toprule
\textbf{Model} & \textbf{Minimal} & \textbf{Band} & \textbf{Retained} & \textbf{Fixed--sham} & \textbf{\(n\)} \\
\midrule
\texttt{deepseek-v4-flash} & 0.6029 & Yes & Yes & 0.1343 & 67 \\
\texttt{doubao-pro} & 0.5217 & Yes & Yes & 0.0435 & 69 \\
\texttt{kimi-32k} & 0.1159 & No & Yes & -0.0145 & 69 \\
\texttt{minimax-text} & 0.0 & No & No & 0.0 & 27 \\
\texttt{qwen-turbo} & 0.1017 & No & Yes & 0.1333 & 60 \\
\bottomrule
\end{tabular}
\end{adjustbox}
\end{table*}

Table~\ref{tab:app_planner_headroom} records the Minimal success rate
and pooled-retention flag, defined by a rate of at least 0.05.
The intermediate band denotes baseline success in [0.40,0.80], an
exploratory display category. Baseline success is a
model attribute, not a validated threshold for activating planning.

\begin{table*}[!htbp]
\centering
\small
\setlength{\tabcolsep}{4pt}
\caption{Exploratory task-complexity moderation.}
\label{tab:app_planner_complexity}
\begin{adjustbox}{max width=\linewidth,max totalheight=0.82\textheight}
\begin{tabular}{lrrrrrr}
\toprule
\textbf{Feature} & \textbf{Median} & \textbf{Low effect} & \textbf{\(n_L\)} & \textbf{High effect} & \textbf{\(n_H\)} & \textbf{Slope} \\
\midrule
\texttt{x\_substeps} & 4.0 & 0.0211 & 142 & 0.1301 & 123 & 0.0178 \\
\texttt{x\_dependency\_depth} & 3.0 & 0.0211 & 142 & 0.1301 & 123 & 0.0172 \\
\texttt{x\_tool\_count} & 3.0 & 0.0211 & 142 & 0.1301 & 123 & 0.0318 \\
\bottomrule
\end{tabular}
\end{adjustbox}
\end{table*}

Low- and high-complexity effects are 0.0211 and 0.1301.
The continuous slopes summarize these correlated
reference-action features.

\subsection{Task Inclusion and Replication Sensitivity}

The exposure-restricted sensitivity retains primary pairs only when both
archived injection records contain nonempty text. The remaining 240 pairs
contain 110 Fixed and 92 Sham successes, giving \((110-92)/240=0.0750\).
Its 90\% task-clustered percentile interval is [0.0091,0.1423], using
5,000 resamples and seed 2701. This is an exposure-selected sensitivity;
the assigned-configuration comparison remains primary.

\begin{table*}[!htbp]
\centering
\small
\setlength{\tabcolsep}{4pt}
\caption{Planning sensitivity to task inclusion and replication. The separate power simulation is described below.}
\label{tab:app_planner_robustness}
\begin{adjustbox}{max width=\linewidth,max totalheight=0.82\textheight}
\begin{tabular}{lrr}
\toprule
\textbf{Check} & \textbf{Point} & \textbf{\(n\)} \\
\midrule
\texttt{mutation\_only\_pooled} & 0.0717 & 265 \\
\texttt{all\_tasks\_pooled} & 0.0758 & 277 \\
\texttt{pooled\_rep=0} & 0.0562 & 89 \\
\texttt{pooled\_rep=1} & 0.0682 & 88 \\
\texttt{pooled\_rep=2} & 0.0909 & 88 \\
\bottomrule
\end{tabular}
\end{adjustbox}
\end{table*}

The mutation-only retained result is \(0.0717\) on 265 pairs; the all-task
result is \(0.0758\) on 277 pairs. Replication-specific estimates are
\(0.0562\), \(0.0682\), and \(0.0909\). These estimates reuse tasks
and should not be treated as independent task-level replications.

A separate two-arm design simulation reports rejection rates of 0.14
for 23 tasks and 0.18 for a hypothetical
46 tasks. It uses three replications, baseline probability 0.40, a logit
shift calibrated near an effect of 0.07, task-effect SD 0.7, and 200
simulations with 300 task-bootstrap draws and 95\% intervals each.
The simulated design uses a single model with 95\% intervals.
The primary analysis pools four models (265 pairs) and uses 90\%
intervals; its design differs from the simulation in both respects.

\subsection{Effort, Arm-Level Outcomes, and Rule-Based Failure Labels}

The following tables report four-model arm-level effort and quality.
Their means use each arm's available trajectories; matched
contrasts use the paired counts specified separately.

\begin{table*}[!htbp]
\centering
\small
\setlength{\tabcolsep}{4pt}
\caption{Planner-focused Retail effort by model and arm.}
\label{tab:app_planner_effort}
\begin{adjustbox}{max width=\linewidth,max totalheight=0.82\textheight}
\begin{tabular}{llrrrr}
\toprule
\textbf{Model} & \textbf{Arm} & \textbf{Tools} & \textbf{Latency (s)} & \textbf{Tokens} & \textbf{Plan tokens} \\
\midrule
\texttt{deepseek-v4-flash} & \texttt{minimal} & 7.84 & 290.99 & 89101.19 & 0.0 \\
\texttt{deepseek-v4-flash} & \texttt{planner\_fixed} & 7.76 & 60.64 & 87367.94 & 0.0 \\
\texttt{deepseek-v4-flash} & \texttt{planner\_self} & 7.44 & 310.77 & 101509.19 & 1148.04 \\
\texttt{deepseek-v4-flash} & \texttt{planner\_sham} & 7.43 & 288.58 & 91942.4 & 0.0 \\
\texttt{doubao-pro} & \texttt{minimal} & 7.23 & 175.91 & 99282.54 & 0.0 \\
\texttt{doubao-pro} & \texttt{planner\_fixed} & 7.06 & 120.2 & 86354.23 & 0.0 \\
\texttt{doubao-pro} & \texttt{planner\_self} & 7.83 & 189.62 & 104787.86 & 2562.06 \\
\texttt{doubao-pro} & \texttt{planner\_sham} & 7.2 & 139.61 & 91550.86 & 0.0 \\
\texttt{kimi-32k} & \texttt{minimal} & 5.52 & 60.45 & 47958.38 & 0.0 \\
\texttt{kimi-32k} & \texttt{planner\_fixed} & 6.72 & 45.64 & 41422.91 & 0.0 \\
\texttt{kimi-32k} & \texttt{planner\_self} & 8.36 & 65.69 & 65512.35 & 678.49 \\
\texttt{kimi-32k} & \texttt{planner\_sham} & 5.8 & 54.71 & 48479.03 & 0.0 \\
\texttt{qwen-turbo} & \texttt{minimal} & 6.83 & 80.69 & 112276.56 & 0.0 \\
\texttt{qwen-turbo} & \texttt{planner\_fixed} & 7.13 & 64.14 & 109396.78 & 0.0 \\
\texttt{qwen-turbo} & \texttt{planner\_self} & 8.15 & 183.92 & 143316.31 & 673.8 \\
\texttt{qwen-turbo} & \texttt{planner\_sham} & 5.73 & 130.86 & 120140.77 & 0.0 \\
\bottomrule
\end{tabular}
\end{adjustbox}
\end{table*}

\begin{table*}[!htbp]
\centering
\small
\setlength{\tabcolsep}{3pt}
\caption{Planner-arm quality outcomes in the four-model detailed analysis.}
\label{tab:app_planner_quality}

\begin{adjustbox}{max width=\linewidth,max totalheight=0.82\textheight}
\begin{tabular}{llrrrr}
\toprule
\textbf{Model} & \textbf{Arm} & \textbf{\(n\)} & \textbf{Verified} & \textbf{False pass} & \textbf{Asserted} \\
\midrule
\texttt{deepseek-v4-flash} & \texttt{minimal} & 68 & 0.6029 & 0.3824 & 0.9853 \\
\texttt{deepseek-v4-flash} & \texttt{planner\_fixed} & 68 & 0.7059 & 0.2794 & 0.9853 \\
\texttt{deepseek-v4-flash} & \texttt{planner\_self} & 68 & 0.6029 & 0.3529 & 0.9559 \\
\texttt{deepseek-v4-flash} & \texttt{planner\_sham} & 68 & 0.5735 & 0.3971 & 0.9706 \\
\texttt{doubao-pro} & \texttt{minimal} & 69 & 0.5217 & 0.4638 & 0.9855 \\
\texttt{doubao-pro} & \texttt{planner\_fixed} & 69 & 0.6377 & 0.3623 & 1.0 \\
\texttt{doubao-pro} & \texttt{planner\_self} & 69 & 0.5942 & 0.3913 & 0.9855 \\
\texttt{doubao-pro} & \texttt{planner\_sham} & 69 & 0.5942 & 0.4058 & 1.0 \\
\texttt{kimi-32k} & \texttt{minimal} & 69 & 0.1159 & 0.8841 & 1.0 \\
\texttt{kimi-32k} & \texttt{planner\_fixed} & 69 & 0.1739 & 0.6667 & 0.8406 \\
\texttt{kimi-32k} & \texttt{planner\_self} & 69 & 0.1159 & 0.6957 & 0.8116 \\
\texttt{kimi-32k} & \texttt{planner\_sham} & 69 & 0.1884 & 0.7681 & 0.9565 \\
\texttt{qwen-turbo} & \texttt{minimal} & 59 & 0.1017 & 0.8305 & 0.9322 \\
\texttt{qwen-turbo} & \texttt{planner\_fixed} & 60 & 0.2 & 0.7667 & 0.9667 \\
\texttt{qwen-turbo} & \texttt{planner\_self} & 59 & 0.2373 & 0.6949 & 0.9322 \\
\texttt{qwen-turbo} & \texttt{planner\_sham} & 60 & 0.0667 & 0.8667 & 0.9333 \\
\bottomrule
\end{tabular}
\end{adjustbox}
\end{table*}

\begin{table*}[!htbp]
\centering
\small
\setlength{\tabcolsep}{4pt}
\caption{Rule-based outcome composition by planning arm. Shares use all trajectories; None denotes oracle success.}
\label{tab:app_planner_failure_types}
\begin{adjustbox}{max width=\linewidth,max totalheight=0.82\textheight}
\begin{tabular}{lrrrr}
\toprule
\textbf{Arm} & \textbf{Budget} & \textbf{Execution} & \textbf{None} & \textbf{Pseudo-completion} \\
\midrule
\texttt{minimal} & 0.0075 & 0.0151 & 0.3434 & 0.634 \\
\texttt{planner\_fixed} & 0.0075 & 0.0451 & 0.4361 & 0.5113 \\
\texttt{planner\_self} & 0.0189 & 0.0604 & 0.3925 & 0.5283 \\
\texttt{planner\_sham} & 0.015 & 0.0188 & 0.3647 & 0.6015 \\
\bottomrule
\end{tabular}
\end{adjustbox}
\end{table*}

Pseudo-completion is the largest rule-based category in each planning
arm. It is assigned when false pass equals one; \texttt{None} denotes
oracle success. The other categories distinguish budget exhaustion and
execution failure.

\begin{table*}[!htbp]
\centering
\small
\setlength{\tabcolsep}{4pt}
\caption{Fixed-minus-sham effects on false-pass probability.}
\label{tab:app_planner_fp}
\begin{adjustbox}{max width=\linewidth,max totalheight=0.82\textheight}
\begin{tabular}{lrrrrr}
\toprule
\textbf{Model} & \textbf{\(n\)} & \textbf{Effect} & \textbf{Lo} & \textbf{Hi} & \textbf{\(p\)} \\
\midrule
\texttt{deepseek-v4-flash} & 67 & -0.1194 & -0.2239 & -0.0149 & 0.0872 \\
\texttt{doubao-pro} & 69 & -0.0435 & -0.1594 & 0.0725 & 0.616 \\
\texttt{kimi-32k} & 69 & -0.1014 & -0.2029 & 0.0 & 0.1176 \\
\texttt{qwen-turbo} & 60 & -0.1 & -0.2 & 0.0 & 0.1568 \\
\bottomrule
\end{tabular}
\end{adjustbox}
\end{table*}

All four Fixed--Sham false-pass point estimates are negative.
Table~\ref{tab:app_planner_fp} reports their task-clustered intervals
and unadjusted bootstrap quantities.

\subsection{Supplementary Shared-Retail Evidence: E03B}

E03B reuses shared Retail with its own task coverage and matched samples.
Table~\ref{tab:app_planner_e03b} gives 90\% task-clustered intervals
and unadjusted bootstrap quantities.

\begin{table*}[!htbp]
\centering
\small
\setlength{\tabcolsep}{4pt}
\caption{Six-model fixed-minus-sham results from the shared Retail block.}
\label{tab:app_planner_e03b}
\begin{adjustbox}{max width=\linewidth,max totalheight=0.82\textheight}
\begin{tabular}{lrrrrrr}
\toprule
\textbf{Model} & \textbf{Minimal} & \textbf{\(n\)} & \textbf{Effect} & \textbf{90\% Lo} & \textbf{90\% Hi} & \textbf{Boot \(p\)} \\
\midrule
\texttt{claude-haiku} & 0.0 & 10 & 0.2 & -0.1667 & 0.7 & 0.522 \\
\texttt{deepseek-v4-flash} & 0.6667 & 43 & 0.1628 & 0.0 & 0.3333 & 0.1432 \\
\texttt{deepseek-v4-pro} & 0.6444 & 45 & 0.1333 & -0.0444 & 0.3111 & 0.2404 \\
\texttt{glm-4-air} & 0.4444 & 45 & 0.2 & 0.0667 & 0.3333 & 0.012 \\
\texttt{glm-4-flash} & 0.0444 & 45 & -0.0222 & -0.0667 & 0.0 & 0.7136 \\
\texttt{qwen-turbo} & 0.0625 & 30 & 0.0667 & -0.1667 & 0.2667 & 0.6952 \\
\bottomrule
\end{tabular}
\end{adjustbox}
\end{table*}

\FloatBarrier

\section{Additional Verification Results}
\label{app:verifier_results}

\subsection{Observed Rejections and Matched Retail Contrasts}

Table~\ref{tab:app_verifier_outcome_decomposition} in the main text
reports the direct rejection counts. The pooled configuration means below
complement those counts and the model-specific matched contrasts.
\begin{table}[t]
    \centering
    \small
    \caption{
    Pooled Retail outcomes for Minimal and verifier-only.
    Differences are verifier-only minus Minimal and are computed from the
    configuration means. Configuration means use their respective samples.
    }
    \label{tab:verifier_retail_main}
    \setlength{\tabcolsep}{6pt}
    \begin{tabular}{@{}lrrr@{}}
        \toprule
        \textbf{Configuration}
        & \textbf{\(n\)}
        & \textbf{Verified success}
        & \textbf{False pass} \\
        \midrule
        \texttt{minimal}
        & 235 & 0.3787 & 0.5830 \\
        \texttt{verifier\_only}
        & 229 & 0.4017 & 0.2096 \\
        Difference
        & -- & 0.0230 & -0.3734 \\
        \bottomrule
    \end{tabular}
\end{table}

Pooled means use all available trajectories in each configuration;
the table's sample sizes identify their coverage.

\begin{table}[!htbp]
\centering\small
\caption{Matched Retail Verifier--Minimal false-pass differences.
90\% task-clustered percentile intervals, 5,000 draws; tests are unadjusted.
\(0^*\) denotes an empty bootstrap tail.}
\label{tab:app_retail_verifier_matched}
\begin{tabular}{@{}lrrrrr@{}}
\toprule Model & Pairs & Difference & Lower & Upper & \(p\)\\\midrule
\texttt{claude-haiku} & 10 & -1.0000 & -1.0000 & -1.0000 & \(0^*\)\\
\texttt{deepseek-v4-flash} & 42 & -0.0952 & -0.2619 & 0.0476 & 0.3744\\
\texttt{deepseek-v4-pro} & 48 & -0.0625 & -0.1667 & 0.0417 & 0.4068\\
\texttt{glm-4-air} & 48 & -0.3750 & -0.5208 & -0.2490 & \(0^*\)\\
\texttt{glm-4-flash} & 48 & -0.5625 & -0.7083 & -0.3958 & \(0^*\)\\
\texttt{qwen-turbo} & 33 & -0.6970 & -0.8788 & -0.4848 & \(0^*\)\\
\bottomrule\end{tabular}
\end{table}
Claude's ten differences are all negative one, producing a
degenerate resampling interval. Precision is limited by this small
support. The other models also have negative point estimates,
with interval widths reflecting task variation and paired coverage.

\subsection{FDS-by-Verifier Interaction}

The capability-conditioned analysis uses 1,547 shared Retail trajectories
and six model clusters. Table~\ref{tab:app_verifier_interaction} retains
the coefficient, clustered inference, and small-sample diagnostics.

\begin{table*}[!htbp]
\centering
\small
\setlength{\tabcolsep}{4pt}
\caption{FDS-by-Verifier interaction and small-cluster diagnostics. Conventional clustered inference is shown for completeness; the delete-one-model jackknife motivates the conservative interpretation in the text.}
\label{tab:app_verifier_interaction}
\begin{adjustbox}{max width=\linewidth,max totalheight=0.82\textheight}
\begin{tabular}{lrrrrrrr}
\toprule
\textbf{Term} & \textbf{Coef.} & \textbf{Cluster SE} & \textbf{Cluster \(p\)} & \textbf{LOMO min} & \textbf{LOMO max} & \textbf{Jackknife SE} & \textbf{Jackknife \(p\)} \\
\midrule
\texttt{FDSz:h\_verifier} & 0.1825 & 0.0460 & 7.2e-05 & 0.1492 & 0.5078 & 0.2803 & 0.5439 \\
\bottomrule
\end{tabular}
\end{adjustbox}
\end{table*}

\begin{table}[!htbp]
\centering
\small
\setlength{\tabcolsep}{4pt}
\caption{Leave-one-model-out FDS-by-Verifier coefficients.}
\label{tab:app_verifier_lomo}
\begin{adjustbox}{max width=\linewidth,max totalheight=0.82\textheight}
\begin{tabular}{lr}
\toprule
\textbf{Dropped model} & \textbf{Coefficient} \\
\midrule
\texttt{claude-haiku} & 0.1787 \\
\texttt{deepseek-v4-flash} & 0.1752 \\
\texttt{deepseek-v4-pro} & 0.1492 \\
\texttt{glm-4-air} & 0.1903 \\
\texttt{glm-4-flash} & 0.5078 \\
\texttt{qwen-turbo} & 0.1717 \\
\bottomrule
\end{tabular}
\end{adjustbox}
\end{table}

The interaction coefficient is 0.1825, with conventional model-clustered
SE 0.0460 and \(p=7.2\times10^{-5}\). The delete-one-model diagnostics assess uncertainty with only six model clusters.

The positive interaction associates higher measured FDS with a smaller
additional false-pass reduction under verification in these six models.

LOMO coefficients range from 0.1492 to 0.5078. The jackknife SE is
0.2803 and its \(p\)-value is 0.5439. The coefficient's sign is stable
across deletions, while its magnitude and precision remain uncertain.

\FloatBarrier

\section{Complete Liability Scenarios and Break-Even Results}
\label{app:liability_results}

\subsection{Scenario Grid, Risk Labels, and Analysis Samples}

Risk is assigned from reference action names as described in
Section~\ref{subsec:liability_scenario_inputs}. All low-risk rows
come from task 62. At each risk/configuration combination, differences
are first averaged over matched task--replication pairs within each model.
The scenario summary then averages the model-level values equally over
models and \(V\in\{1,5,20\}\), separately for \(L=2\) and \(L=200\).
The intermediate model differences are recorded to four decimals for
success and false pass and six for cost in the frozen pipeline.

Table~\ref{tab:app_e05_pair_coverage}
reports the underlying paired coverage independently of this grid.
\begin{table}[!htbp]\centering\small
\caption{E05 matched-pair coverage before expanding the value grid.
Low risk contains only task 62.}
\label{tab:app_e05_pair_coverage}
\begin{tabular}{@{}lrrrr@{}}
\toprule & \multicolumn{2}{c}{High risk} & \multicolumn{2}{c}{Low risk}\\
Configuration & Models & Pairs & Models & Pairs\\\midrule
Evaluator-only & 6 & 217 & 5 & 15\\
Full Fixed & 6 & 135 & 3 & 7\\
Orchestrator-only & 6 & 215 & 5 & 15\\
Planner Fixed & 6 & 217 & 5 & 15\\
Planner Sham & 6 & 218 & 5 & 15\\
Verifier-only & 6 & 214 & 5 & 15\\
\bottomrule\end{tabular}\end{table}

\subsection{Complete Net-Value Table}

Table~\ref{tab:app_liability_full} reports all 24 aggregate scenario
values and their component differences. Values are calculated from the
frozen pipeline inputs and rounded for display.

\begin{table*}[!htbp]
\centering
\small
\setlength{\tabcolsep}{3pt}
\caption{Complete E05 scenario outputs. Means weight contributing models and three task-value settings equally; n counts model--value rows.}
\label{tab:app_liability_full}

\begin{adjustbox}{max width=\linewidth,max totalheight=0.82\textheight}
\begin{tabular}{lllrrrrr}
\toprule
\textbf{Config} & \textbf{Risk} & \textbf{\(L\) (USD)} & \textbf{Mean NV} & \textbf{\(\Delta P\)} & \textbf{Avoided FP} & \textbf{\(\Delta C\)} & \textbf{Scenario rows} \\
\midrule
\texttt{evaluator\_only} & \texttt{high} & 200 & 13.1945 & 0.0428 & 0.0643 & 0.0429 & 18 \\
\texttt{evaluator\_only} & \texttt{high} & 2 & 0.4565 & 0.0428 & 0.0643 & 0.0429 & 18 \\
\texttt{evaluator\_only} & \texttt{low} & 200 & 41.1508 & 0.1333 & 0.2 & 0.0047 & 15 \\
\texttt{evaluator\_only} & \texttt{low} & 2 & 1.5508 & 0.1333 & 0.2 & 0.0047 & 15 \\
\texttt{full\_fixed} & \texttt{high} & 200 & 99.1962 & -0.0411 & 0.4982 & 0.0947 & 18 \\
\texttt{full\_fixed} & \texttt{high} & 2 & 0.5460 & -0.0411 & 0.4982 & 0.0947 & 18 \\
\texttt{full\_fixed} & \texttt{low} & 200 & 92.7301 & 0.4444 & 0.4444 & 0.0083 & 9 \\
\texttt{full\_fixed} & \texttt{low} & 2 & 4.7323 & 0.4444 & 0.4444 & 0.0083 & 9 \\
\texttt{orchestrator\_only} & \texttt{high} & 200 & 9.4953 & 0.0704 & 0.0445 & 0.0047 & 18 \\
\texttt{orchestrator\_only} & \texttt{high} & 2 & 0.6942 & 0.0704 & 0.0445 & 0.0047 & 18 \\
\texttt{orchestrator\_only} & \texttt{low} & 200 & 41.7296 & 0.2000 & 0.2000 & -0.0004 & 15 \\
\texttt{orchestrator\_only} & \texttt{low} & 2 & 2.1335 & 0.2000 & 0.2000 & -0.0004 & 15 \\
\texttt{planner\_fixed} & \texttt{high} & 200 & 35.1299 & 0.1897 & 0.1675 & 0.0041 & 18 \\
\texttt{planner\_fixed} & \texttt{high} & 2 & 1.9748 & 0.1897 & 0.1675 & 0.0041 & 18 \\
\texttt{planner\_fixed} & \texttt{low} & 200 & 13.9092 & 0.0667 & 0.0667 & 0.0006 & 15 \\
\texttt{planner\_fixed} & \texttt{low} & 2 & 0.7105 & 0.0667 & 0.0667 & 0.0006 & 15 \\
\texttt{planner\_sham} & \texttt{high} & 200 & 9.1419 & 0.0580 & 0.0432 & 0.0044 & 18 \\
\texttt{planner\_sham} & \texttt{high} & 2 & 0.5850 & 0.0580 & 0.0432 & 0.0044 & 18 \\
\texttt{planner\_sham} & \texttt{low} & 200 & 27.8198 & 0.1333 & 0.1333 & -0.0004 & 15 \\
\texttt{planner\_sham} & \texttt{low} & 2 & 1.4225 & 0.1333 & 0.1333 & -0.0004 & 15 \\
\texttt{verifier\_only} & \texttt{high} & 200 & 97.3133 & 0.0655 & 0.4838 & 0.0079 & 18 \\
\texttt{verifier\_only} & \texttt{high} & 2 & 1.5275 & 0.0655 & 0.4838 & 0.0079 & 18 \\
\texttt{verifier\_only} & \texttt{low} & 200 & 27.2466 & 0.0667 & 0.1333 & -0.0009 & 15 \\
\texttt{verifier\_only} & \texttt{low} & 2 & 0.8453 & 0.0667 & 0.1333 & -0.0009 & 15 \\
\bottomrule
\end{tabular}
\end{adjustbox}
\end{table*}

\subsection{Complete Break-Even Results}

Table~\ref{tab:app_break_even_full} applies
Equation~\ref{eq:framework_break_even} to the aggregate mean differences
for each configuration and risk stratum.

\begin{table*}[!htbp]
\centering
\small
\setlength{\tabcolsep}{4pt}
\caption{Aggregate break-even false-pass liability from the frozen mean deltas.}
\label{tab:app_break_even_full}
\begin{adjustbox}{max width=\linewidth,max totalheight=0.82\textheight}
\begin{tabular}{llrrrrrr}
\toprule
\textbf{Config} & \textbf{Risk} & \textbf{\(\Delta P\)} & \textbf{Avoided FP} & \textbf{\(\Delta C\)} & \textbf{\(L^*(1)\)} & \textbf{\(L^*(5)\)} & \textbf{\(L^*(20)\)} \\
\midrule
\texttt{evaluator\_only} & high & 0.0428 & 0.0643 & 0.0429 & 0.0021 & -2.6580 & -12.6333 \\
\texttt{evaluator\_only} & low & 0.1333 & 0.2000 & 0.0047 & -0.6432 & -3.3096 & -13.3086 \\
\texttt{full\_fixed} & high & -0.0411 & 0.4982 & 0.0947 & 0.2724 & 0.6020 & 1.8378 \\
\texttt{full\_fixed} & low & 0.4444 & 0.4444 & 0.0083 & -0.9813 & -4.9813 & -19.9813 \\
\texttt{orchestrator\_only} & high & 0.0704 & 0.0445 & 0.0047 & -1.4774 & -7.8110 & -31.5623 \\
\texttt{orchestrator\_only} & low & 0.2000 & 0.2000 & -0.0004 & -1.0020 & -5.0020 & -20.0020 \\
\texttt{planner\_fixed} & high & 0.1897 & 0.1675 & 0.0041 & -1.1083 & -5.6398 & -22.6329 \\
\texttt{planner\_fixed} & low & 0.0667 & 0.0667 & 0.0006 & -0.9914 & -4.9914 & -19.9914 \\
\texttt{planner\_sham} & high & 0.0580 & 0.0432 & 0.0044 & -1.2417 & -6.6130 & -26.7555 \\
\texttt{planner\_sham} & low & 0.1333 & 0.1333 & -0.0004 & -1.0030 & -5.0030 & -20.0030 \\
\texttt{verifier\_only} & high & 0.0655 & 0.4838 & 0.0079 & -0.1192 & -0.6609 & -2.6924 \\
\texttt{verifier\_only} & low & 0.0667 & 0.1333 & -0.0009 & -0.5069 & -2.5066 & -10.0054 \\
\bottomrule
\end{tabular}
\end{adjustbox}
\end{table*}

Full Fixed in the high-risk stratum has positive thresholds \(0.2724\),
\(0.6020\), and \(1.8378\) at \(V=1,5,20\). Its negative verified-success
difference and positive incremental cost require a positive value for
avoided false passes. Evaluator-only in the high-risk stratum also has
a positive threshold at \(V=1\), namely \(0.0021\). Negative entries
mean positive aggregate scenario value at zero false-pass liability for
the corresponding task value.

Section~\ref{subsec:liability_sample_interpretation} discusses the grid
and accounting scope.

\FloatBarrier

\section{Incremental Predictive Validity}
\label{app:incremental_validity}

\subsection{Scalar Benchmark Inputs}

Table~\ref{tab:app_bench_scores} reports scalar benchmark scores and
observation counts for six models. Claude's 72-row batch comprises 24
items with three replications; counts refer to observations.

\begin{table}[!htbp]
\centering
\small
\caption{Scalar benchmark scores and observation counts. Observations include replications.}
\label{tab:app_bench_scores}
\begin{adjustbox}{max width=\linewidth,max totalheight=0.82\textheight}
\begin{tabular}{lrr}
\toprule
\textbf{Model}
&
\textbf{\(\mathrm{Bench}_m\)}
&
\textbf{Observations}
\\
\midrule
\texttt{claude-haiku} & 1.0000 & 72 \\
\texttt{deepseek-v4-flash} & 0.9583 & 72 \\
\texttt{deepseek-v4-pro} & 0.9583 & 72 \\
\texttt{glm-4-air} & 0.8750 & 72 \\
\texttt{qwen-turbo} & 0.7917 & 72 \\
\texttt{glm-4-flash} & 0.7083 & 72 \\
\bottomrule
\end{tabular}
\end{adjustbox}
\end{table}

\subsection{Leave-One-Model-Out Prediction}

E06 predicts trajectory-level verified success by OLS. The baseline
includes the scalar benchmark score, reference-action count,
dependency-depth proxy, and distinct tool-name count. Each fold trains
on five models and evaluates the sixth, clipping predictions to [0,1].
We compare two sources of PF/TCC/FDS features.

The frozen negative-result specification derives capability proxies
from Minimal trajectories using task cross-fitting. Within a model,
the feature vector can vary with task fold, and the held-out model's
features use its own Minimal outcomes from other folds. The independent
probe sensitivity instead joins model-level scores from the separate
E01 held-out probe battery, also used in the FDS-by-verifier analysis
(E04). It keeps the same outcome, task controls, OLS fit, and model folds,
changing only the capability-feature source.

\begin{table}[!htbp]\centering\small
\caption{Leave-one-model-out prediction with different capability inputs.
Each specification includes task controls. Values are computed across
six models; lower Brier and higher \(R^2\)/AUC are favorable.}
\label{tab:app_incremental_validity}
\begin{tabular}{@{}lrrr@{}}
\toprule Input specification & \(R^2\) & AUC & Brier\\\midrule
Benchmark baseline & 0.152 & 0.734 & 0.204\\
Benchmark + cross-fitted trajectory proxies & 0.083 & 0.715 & 0.220\\
Benchmark + held-out probes & 0.154 & 0.744 & 0.203\\
\bottomrule\end{tabular}\end{table}

Trajectory proxies worsen all three metrics. Independent probes change
\(R^2\) by +0.002, AUC by +0.010, and Brier by -0.0005, computed before
rounding the displayed scores. Across six model folds, these small changes
do not establish a reliable predictive advantage; uncertainty intervals
were not estimated. The negative result belongs to the trajectory proxies,
not to the independent probes.

\subsection{Claude Benchmark Retry and Source Discrepancy}

A 72-row Claude batch failed with an account-balance error. Its successful
72-row retry supplies the analysis input, with all rows valid and mean
hit rate 1.0. The failed requests remain in the coverage audit and are
excluded from behavioral scoring.

\FloatBarrier

\section{Airline Pilot Results}
\label{app:airline_results}

\subsection{Realized Coverage}

Airline contains 233 of 240 planned trajectories from five models, six
tasks, four configurations, and two replications.
Table~\ref{tab:app_airline_coverage} pools outcomes across configurations
within each model.

\begin{table*}[!htbp]
\centering
\small
\setlength{\tabcolsep}{4pt}
\caption{Airline realized coverage by model.}
\label{tab:app_airline_coverage}
\begin{adjustbox}{max width=\linewidth,max totalheight=0.82\textheight}
\begin{tabular}{lrrrrrr}
\toprule
\textbf{Model} & \textbf{Rows} & \textbf{Tasks} & \textbf{Configs} & \textbf{Reps} & \textbf{Verified} & \textbf{False pass} \\
\midrule
\texttt{claude-haiku} & 43 & 6 & 4 & 2 & 0.2791 & 0.5581 \\
\texttt{deepseek-v4-flash} & 48 & 6 & 4 & 2 & 0.5000 & 0.3958 \\
\texttt{doubao-pro} & 46 & 6 & 4 & 2 & 0.6304 & 0.2826 \\
\texttt{glm-4-air} & 48 & 6 & 4 & 2 & 0.3333 & 0.5000 \\
\texttt{qwen-turbo} & 48 & 6 & 4 & 2 & 0.1458 & 0.6875 \\
\bottomrule
\end{tabular}
\end{adjustbox}
\end{table*}

\subsection{Planning Contrasts}

Table~\ref{tab:app_airline_planner} retains the matched Fixed--Sham
contrasts, bounds, and Holm-adjusted \(p\)-values. The largest
point estimate is \(0.25\) for \texttt{deepseek-v4-flash}; its
Holm-adjusted \(p\)-value is \(0.14\). None of the adjusted
values is below \(0.05\).

\begin{table*}[!htbp]
\centering
\small
\setlength{\tabcolsep}{4pt}
\caption{Airline fixed-minus-sham planner effects.}
\label{tab:app_airline_planner}
\begin{adjustbox}{max width=\linewidth,max totalheight=0.82\textheight}
\begin{tabular}{lrrrrrr}
\toprule
\textbf{Model} & \textbf{\(n\)} & \textbf{Effect} & \textbf{Lo} & \textbf{Hi} & \textbf{\(p\)} & \textbf{Holm \(p\)} \\
\midrule
\texttt{claude-haiku} & 10 & -0.1 & -0.3 & 0.0 & 0.6532 & 1.0 \\
\texttt{deepseek-v4-flash} & 12 & 0.25 & 0.0833 & 0.4167 & 0.028 & 0.14 \\
\texttt{doubao-pro} & 12 & 0.0833 & -0.1667 & 0.3333 & 0.738 & 1.0 \\
\texttt{glm-4-air} & 12 & 0.0 & -0.25 & 0.25 & 1.0 & 1.0 \\
\texttt{qwen-turbo} & 12 & 0.0 & -0.1667 & 0.1667 & 1.0 & 1.0 \\
\bottomrule
\end{tabular}
\end{adjustbox}
\end{table*}

\begin{table*}[!htbp]
\centering
\small
\setlength{\tabcolsep}{3pt}
\caption{Airline secondary planning contrasts with 90\% matched-cell bootstrap intervals. Self-planning was not run.}
\label{tab:app_airline_secondary}

\begin{adjustbox}{max width=\linewidth,max totalheight=0.82\textheight}
\begin{tabular}{llrrrrr}
\toprule
\textbf{Model} & \textbf{Contrast} & \textbf{\(n\)} & \textbf{Effect} & \textbf{Lo} & \textbf{Hi} & \textbf{\(p\)} \\
\midrule
\texttt{claude-haiku} & \texttt{fixed-minimal} & 11 & 0.0909 & -0.1818 & 0.3636 & 0.7688 \\
\texttt{deepseek-v4-flash} & \texttt{fixed-minimal} & 12 & 0.4167 & 0.1667 & 0.6667 & 0.002 \\
\texttt{doubao-pro} & \texttt{fixed-minimal} & 10 & 0.2 & 0.0 & 0.4 & 0.2256 \\
\texttt{glm-4-air} & \texttt{fixed-minimal} & 12 & -0.1667 & -0.4167 & 0.0833 & 0.4348 \\
\texttt{qwen-turbo} & \texttt{fixed-minimal} & 12 & 0.1667 & 0.0 & 0.3333 & 0.22 \\
\bottomrule
\end{tabular}
\end{adjustbox}
\end{table*}

Fixed--Minimal jointly changes guidance and added context. Self-planning
was not run in Airline.

\subsection{Verifier-Minus-Minimal False-Pass Contrasts}

\begin{table*}[!htbp]
\centering
\small
\setlength{\tabcolsep}{4pt}
\caption{Airline verifier-minus-Minimal false-pass effects.}
\label{tab:app_airline_verifier}
\begin{adjustbox}{max width=\linewidth,max totalheight=0.82\textheight}
\begin{tabular}{lrrrrr}
\toprule
\textbf{Model} & \textbf{\(n\)} & \textbf{\(\Delta FP\)} & \textbf{Lo} & \textbf{Hi} & \textbf{\(p\)} \\
\midrule
\texttt{claude-haiku} & 10 & -0.7 & -1.0 & -0.1 & 0.0144 \\
\texttt{deepseek-v4-flash} & 12 & -0.25 & -0.5833 & 0.0833 & 0.2832 \\
\texttt{doubao-pro} & 10 & 0.1 & -0.1818 & 0.4 & 0.756 \\
\texttt{glm-4-air} & 12 & -0.25 & -0.5833 & 0.0833 & 0.2952 \\
\texttt{qwen-turbo} & 12 & -0.9167 & -1.0 & -0.75 & \(0^*\) \\
\bottomrule
\end{tabular}
\end{adjustbox}
\end{table*}

The largest false-pass reductions are for \texttt{claude-haiku}
(-0.7000) and \texttt{qwen-turbo} (-0.9167). Other estimates are smaller
or positive, with 90\% task-clustered intervals spanning zero.

Qwen's \(0^*\) marks an empty bootstrap tail among 5,000 draws,
following the convention in Appendix~\ref{app:statistical_inference}.

\subsection{Seven Missing Planned Cells}

\begin{table}[!htbp]
\centering
\small
\setlength{\tabcolsep}{4pt}
\caption{Seven missing Airline cells.}
\label{tab:app_airline_missing}
\begin{adjustbox}{max width=\linewidth,max totalheight=0.82\textheight}
\begin{tabular}{lllr}
\toprule
\textbf{Model} & \textbf{Task} & \textbf{Config} & \textbf{Rep} \\
\midrule
\texttt{claude-haiku} & 21 & \texttt{planner\_fixed} & 1 \\
\texttt{claude-haiku} & 21 & \texttt{planner\_sham} & 0 \\
\texttt{claude-haiku} & 21 & \texttt{planner\_sham} & 1 \\
\texttt{claude-haiku} & 21 & \texttt{verifier\_only} & 0 \\
\texttt{claude-haiku} & 21 & \texttt{verifier\_only} & 1 \\
\texttt{doubao-pro} & 18 & \texttt{minimal} & 0 \\
\texttt{doubao-pro} & 21 & \texttt{minimal} & 1 \\
\bottomrule
\end{tabular}
\end{adjustbox}
\end{table}

The five missing Claude cells concern task 21 under Fixed, Sham, and
Verifier-only. The two missing Doubao cells are Minimal runs on tasks 18
and 21. These gaps explain part of the variation in matched contrast counts.
Their behavioral outcomes are not imputed, and the source does not establish
that the missingness is random.

\subsection{Cross-Domain Evidence Scope}

\begin{figure}[t]
\centering
\includegraphics[width=0.92\linewidth]{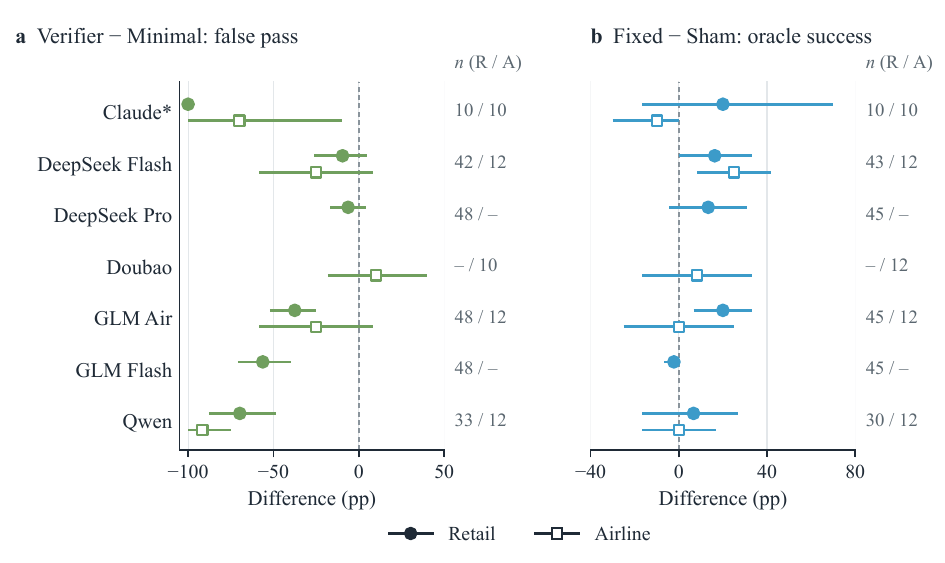}
\caption{\textbf{Retail and Airline model-specific contrasts.}
(a) Verifier--Minimal false-pass differences. (b) Fixed--Sham oracle-success
differences. Filled circles show Retail and hollow squares show Airline.
Retail planning uses E03B; Airline uses the six-task E07 pilot.
Bars are 90\% task-clustered percentile bootstrap intervals (5,000 draws).
The count columns give matched Retail/Airline pairs (R/A); a dash marks
unavailable coverage. *Claude's Retail false-pass contrast has ten identical differences
of \(-1\), giving a degenerate interval at \(-100\) percentage points.
The domain-specific tables report the full numerical estimates.}
\label{fig:app_cross_domain}
\end{figure}

Figure~\ref{fig:app_cross_domain} compares Retail and Airline
model-specific contrasts. Airline contributes six tasks with two
replications, providing a pilot-scale second-environment comparison.

\FloatBarrier

\section{Tool-Coordination Measurement Failure}
\label{app:tcc_failure}

\subsection{Base Instrument and Stress Extension}

The base TCC probe assesses tool selection, schema compliance, preconditions,
action ordering, and error recovery. It contains 162 observations across the
six models in the capability study. The scores show substantial
score concentration near one.

The stress extension adds six items with three attempts per item across
eleven model identifiers, giving 198 attempts. Its coverage is reported
separately from the six-model base instrument.

\begin{table*}[!htbp]
\centering
\small
\setlength{\tabcolsep}{4pt}
\caption{TCC reliability before and after the stress extension. The base estimate is the E01 summary in Table~\ref{tab:app_axis_reliability}; the merged estimate uses the E08 stress analysis.}
\label{tab:app_tcc_reliability}
\begin{adjustbox}{max width=\linewidth,max totalheight=0.82\textheight}
\begin{tabular}{lrrrrl}
\toprule
\textbf{Scope} & \textbf{Rows} & \textbf{Reliability} & \textbf{\(q_{.05}\)} & \textbf{\(q_{.95}\)} & \textbf{0.70 threshold} \\
\midrule
\texttt{E01 base-only} & 162 & -2.116 & -2.379 & -0.690 & Fail \\
\texttt{base + stress merged} & 270 & 0.349 & -2.241 & 0.659 & Fail \\
\bottomrule
\end{tabular}
\end{adjustbox}
\end{table*}

\begin{table*}[!htbp]
\centering
\small
\setlength{\tabcolsep}{4pt}
\caption{TCC stress-probe coverage and mean score.}
\label{tab:app_tcc_coverage}
\begin{adjustbox}{max width=\linewidth,max totalheight=0.82\textheight}
\begin{tabular}{lrrrr}
\toprule
\textbf{Model} & \textbf{Attempts} & \textbf{Valid} & \textbf{Errors} & \textbf{Mean} \\
\midrule
\texttt{claude-haiku} & 18 & 18 & 0 & 1.0000 \\
\texttt{deepseek-v4-flash} & 18 & 18 & 0 & 0.9861 \\
\texttt{deepseek-v4-pro} & 18 & 18 & 0 & 0.9722 \\
\texttt{doubao-pro} & 18 & 18 & 0 & 1.0000 \\
\texttt{glm-4-air} & 18 & 18 & 0 & 1.0000 \\
\texttt{glm-4-flash} & 18 & 18 & 0 & 0.9167 \\
\texttt{glm-4.6} & 18 & 9 & 9 & 1.0000 \\
\texttt{kimi-8k} & 18 & 0 & 18 & -- \\
\texttt{qwen-max} & 18 & 18 & 0 & 1.0000 \\
\texttt{qwen-plus} & 18 & 18 & 0 & 1.0000 \\
\texttt{qwen-turbo} & 18 & 18 & 0 & 1.0000 \\
\bottomrule
\end{tabular}
\end{adjustbox}
\end{table*}

\subsection{Reliability, Coverage, and Saturation}

Base reliability is \(-2.116\). The base-plus-stress merged estimate is
\(0.349\), with bounds \([-2.241,0.659]\), below the specified
\(0.70\) reliability criterion. The merged 270-row reliability sample
differs from the entire eleven-identifier stress collection.

Of 198 stress attempts, 171 are valid and 27 fail. \texttt{kimi-8k} has no
valid responses, while \texttt{glm-4.6} has nine valid responses and nine
errors. Among the six models shared with the main capability experiment,
stress means range from \(0.9167\) to \(1.0000\), indicating continued
saturation with limited discrimination across models.

\subsection{Interpretation of Measurement Failure}

The instrument has limited discrimination of tool coordination in this
sample, with low reliability and concentrated scores. Evaluating a
TCC-by-Orchestrator relationship requires a more reliable measure.

A replacement could test changing states, overlapping tool affordances,
preconditions, recoverable errors, consequential side effects, and adaptive
difficulty. These features provide directions for a revised instrument.

\FloatBarrier

\section{Data Quality, Missingness, and Reproducibility}
\label{app:data_quality}

\subsection{Coverage Audit}

Table~\ref{tab:app_raw_coverage} records raw-data coverage, duplicate keys,
and valid-response totals. E02--E06 share the same Retail trajectories;
the row identifies their common data source.

\begin{table*}[!htbp]
\centering
\small
\setlength{\tabcolsep}{4pt}
\caption{Raw-data coverage audit.}
\label{tab:app_raw_coverage}
\begin{adjustbox}{max width=\linewidth,max totalheight=0.82\textheight}
\begin{tabular}{lrrrrrr}
\toprule
\textbf{Dataset} & \textbf{Rows} & \textbf{Valid} & \textbf{Models} & \textbf{Tasks/configs} & \textbf{Reps} & \textbf{Duplicate keys} \\
\midrule
\texttt{E01 held-out capability probes} & 486 & 486 & 6 & \texttt{27 items / 3 axes} & 3 & 0 \\
\texttt{E02-E06 Retail trajectories} & 1547 & 1547 & 6 & \texttt{16 tasks / 7 configs} & 3 & 0 \\
\texttt{E03A Retail planner trajectories} & 1227 & 1227 & 5 & \texttt{24 tasks / 4 configs} & 3 & 0 \\
\texttt{E06 Claude Bench\_m failed batch} & 72 & 0 & 1 & \texttt{24 items} & 3 & 0 \\
\texttt{E06 Claude Bench\_m successful retry} & 72 & 72 & 1 & \texttt{24 items} & 3 & 0 \\
\texttt{E07 Airline trajectories} & 233 & 233 & 5 & \texttt{6 tasks / 4 configs} & 2 & 0 \\
\texttt{E08 TCC stress probe} & 198 & 171 & 11 & \texttt{6 items} & 3 & 0 \\
\bottomrule
\end{tabular}
\end{adjustbox}
\end{table*}

E01 contributes 486 probe responses and model-level scores to E04 and
the held-out-probe version of E06. The shared trajectory analyses use
1,547 Retail episodes. The failed and successful Claude benchmark batches
are recorded separately; only the successful batch enters E06 scoring.

\subsection{Planner Coverage and Retention}

\begin{table}[!htbp]
\centering
\small
\setlength{\tabcolsep}{4pt}
\caption{E03A model-by-arm realized row counts.}
\label{tab:app_e03_counts}
\begin{adjustbox}{max width=\linewidth,max totalheight=0.82\textheight}
\begin{tabular}{lrrrr}
\toprule
\textbf{Model} & \textbf{Minimal} & \textbf{Fixed} & \textbf{Self} & \textbf{Sham} \\
\midrule
\texttt{deepseek-v4-flash} & 71 & 71 & 71 & 71 \\
\texttt{doubao-pro} & 72 & 72 & 72 & 72 \\
\texttt{kimi-32k} & 72 & 72 & 72 & 72 \\
\texttt{minimax-text} & 30 & 29 & 29 & 29 \\
\texttt{qwen-turbo} & 62 & 63 & 62 & 63 \\
\bottomrule
\end{tabular}
\end{adjustbox}
\end{table}

The frozen planner data contain five identifiers and 1,227 trajectories.
Relative to the complete \(5\times24\times4\times3\) design, 213 cells
are absent. These absent cells remain outside behavioral scoring.

Detailed effort and quality tables use four retained models, the
model-specific Fixed--Sham table includes all five, and the primary pooled
result uses four models and 23 tasks. These are different analysis scopes.
Sample definitions follow Section~\ref{subsec:design_samples}.

\subsection{Missing Runs, Infrastructure Failures, and Retries}

The seven Airline gaps are enumerated in
Table~\ref{tab:app_airline_missing}. E08 records 27 invocation failures.
The E06 Claude account-balance failure and retry are described in
Appendix~\ref{app:incremental_validity}.

\subsection{Outcome and Cost Documentation Limits}

The scoring definitions, plan exposure, matched verifier results,
Pass@Budget denominator, and E05 risk/aggregation rules are documented
in Sections~\ref{subsec:design_recorded_measures}--\ref{subsec:design_inference_scope}
and Appendices~\ref{app:harness_details}--\ref{app:statistical_inference}.
The outcome cross-tabulation distinguishes oracle-correct rejection
from the historical residual. Zero logged costs and infinite
Cost@Success entries remain telemetry conventions.

Runtime metadata are listed in Appendix~\ref{app:model_metadata};
plan construction is described in Section~\ref{subsec:design_planning_conditions}.

\subsection{Available Materials and Reproducibility Limits}

The research archive contains frozen trajectory tables, analysis scripts,
task-indexed plans, figure source tables, and batch diagnostics. These
materials support the definitions, matching rules, and calculations
reported here. Their public distribution status is described in the
data/code availability statement.

External reproduction also depends on the benchmark task version,
provider access, and hosted model behavior at execution time.
Endpoint drift, incomplete coverage, possible familiarity with public
tasks, and the remaining runtime metadata gaps delimit reproducibility.

\paragraph{Data and code availability.}
The study uses public $\tau^2$-bench environments. The research archive
contains the study trajectories, analysis scripts, plans, and diagnostics;
a public repository link is not yet available at the time of this version.

\end{document}